\documentclass[table]{ai2style/ai2}

\usepackage{microtype}
\usepackage{hyperref}
\usepackage{url}
\usepackage{booktabs, tabularx} 
\usepackage{graphicx}
\usepackage{lineno}
\usepackage{enumitem}
\usepackage{listings} 
\usepackage{svg}

\definecolor{ darkblue}{rgb}{0, 0, 0.5}
\hypersetup{colorlinks=true, citecolor=darkblue, linkcolor=darkblue, urlcolor=darkblue}

\usepackage{amssymb}
\usepackage{multirow}
\usepackage{bigdelim}
\usepackage{todonotes}
\usepackage{longtable}
\usepackage{tabularray}
\usepackage{wrapfig}
\usepackage[most]{tcolorbox}
\usepackage{url}
\usepackage{xspace}
\usepackage{svg}
\usepackage[absolute]{textpos} 

\usepackage{fdsymbol}   

\usepackage[utf8]{inputenc} 
\usepackage[T1]{fontenc}    
\usepackage{url}            
\usepackage{booktabs}       
\usepackage{amsfonts}       
\usepackage{nicefrac}       
\usepackage{microtype}      
\usepackage{amsmath}
\usepackage[most]{tcolorbox}
\usepackage{csquotes}

\usepackage{siunitx}
\usepackage{graphicx}
\usepackage{arydshln}
\usepackage{wrapfig}
\usepackage{enumitem}
\usepackage{soul} 

\usepackage{multirow}
\usepackage{xspace}
\usepackage{adjustbox}
\usepackage{pifont}
\usepackage{caption}
\usepackage{makecell}
\usepackage{subcaption}
\usepackage{bold-extra}
\usepackage{url}

\usepackage{pgf-pie}

\usepackage{hyperref}
\definecolor{linkcolor}{RGB}{0, 0, 128}
\hypersetup{
     colorlinks   = true,
     citecolor    = linkcolor,
     linkcolor    = linkcolor,
     urlcolor     = linkcolor,
}
\usepackage{pifont}
\usepackage{listings}

\setlist[itemize]{leftmargin=*,itemsep=0em,parsep=0.3em,topsep=0.3em}

\DeclareUnicodeCharacter{2212}{\ensuremath{-}}

\usepackage{tikz}

\usepackage{setspace}

\usepackage{nicematrix}
\newcolumntype{L}[1]{>{\raggedright\let\newline\\\arraybackslash\hspace{0pt}}m{#1}}
\newcolumntype{C}[1]{>{\centering\let\newline\\\arraybackslash\hspace{0pt}}m{#1}}
\newcolumntype{R}[1]{>{\raggedleft\let\newline\\\arraybackslash\hspace{0pt}}m{#1}}
\newcolumntype{P}[1]{>{\centering\let\newline\\\arraybackslash\columncolor{ai2lightpink}}m{#1}}
\newcommand{\aitoo}{\raisebox{-1.5pt}{\includegraphics[height=1.05em]{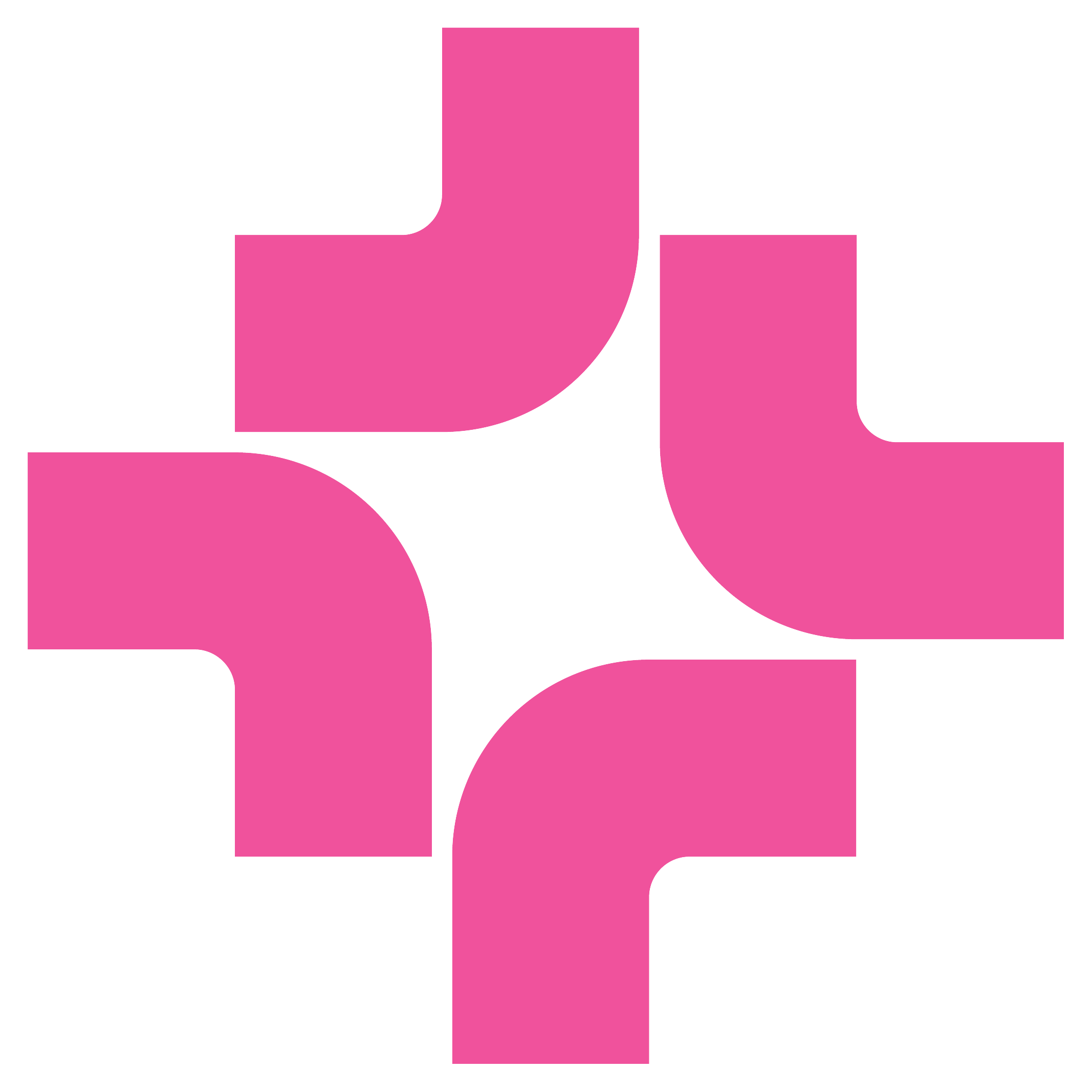}}\xspace}
\newcommand{\allenAiAff}{\raisebox{.28em}{\hspace{.02em}\scalebox{0.7}{\textbf{1}}}}

\newcommand{\uwAff}{\raisebox{.28em}{\hspace{.02em}\scalebox{0.7}{\textbf{2}}}}
\newcommand{\asuAff}{\raisebox{.28em}{\hspace{.02em}\scalebox{0.7}{\textbf{3}}}}
\newcommand{\ubcAff}{\raisebox{.28em}{\hspace{.02em}\scalebox{0.7}{\textbf{4}}}}

\newcommand{\commaAff}{\raisebox{.28em}{\hspace{.02em}\scalebox{0.7}{\textbf{,}\hspace{0.1em}}}}
\newcommand{\coreContrib}{\raisebox{.28em}{\hspace{.05em}\includegraphics[height=.45em]{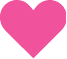}}\hspace{0.1em}}
\newcommand{\starOlmo}{\raisebox{.28em}{\hspace{.05em}\includegraphics[height=.5em]{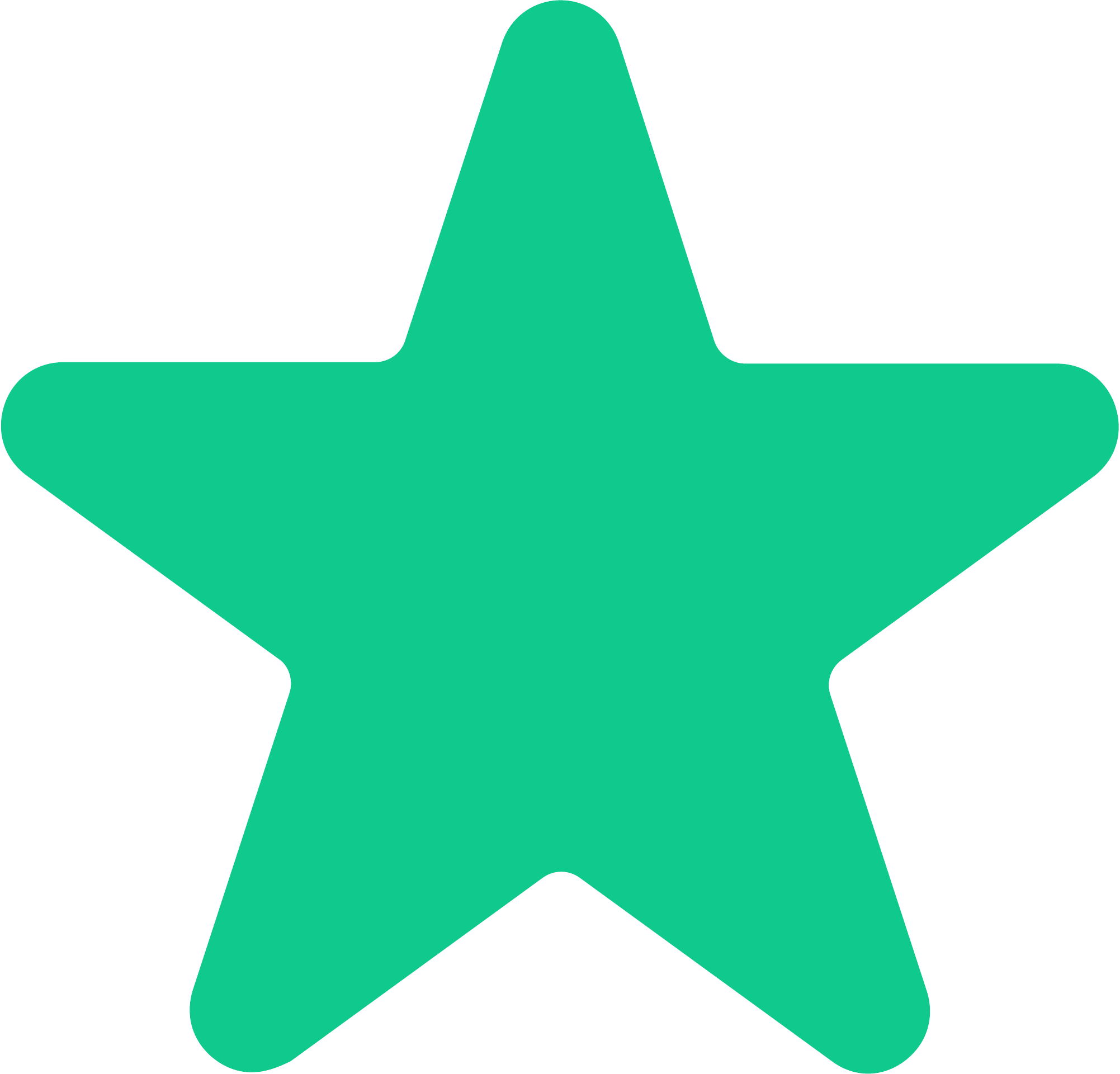}}}
\newcommand{\huggingface}{\raisebox{-1.5pt}{\includegraphics[height=1.05em]{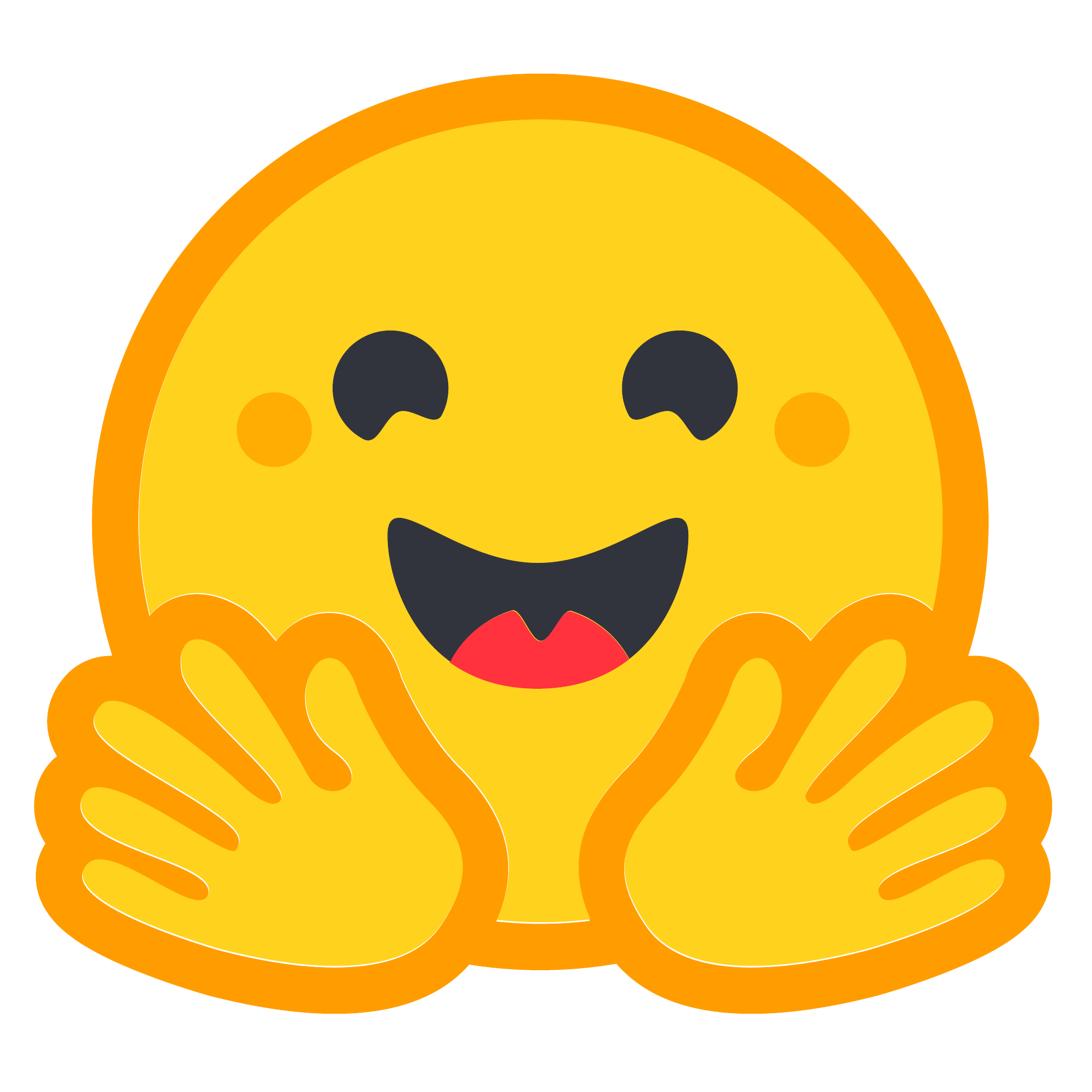}}\xspace}

\newcommand{\emailLogo}{\raisebox{-1.5pt}{\includegraphics[height=1.05em]{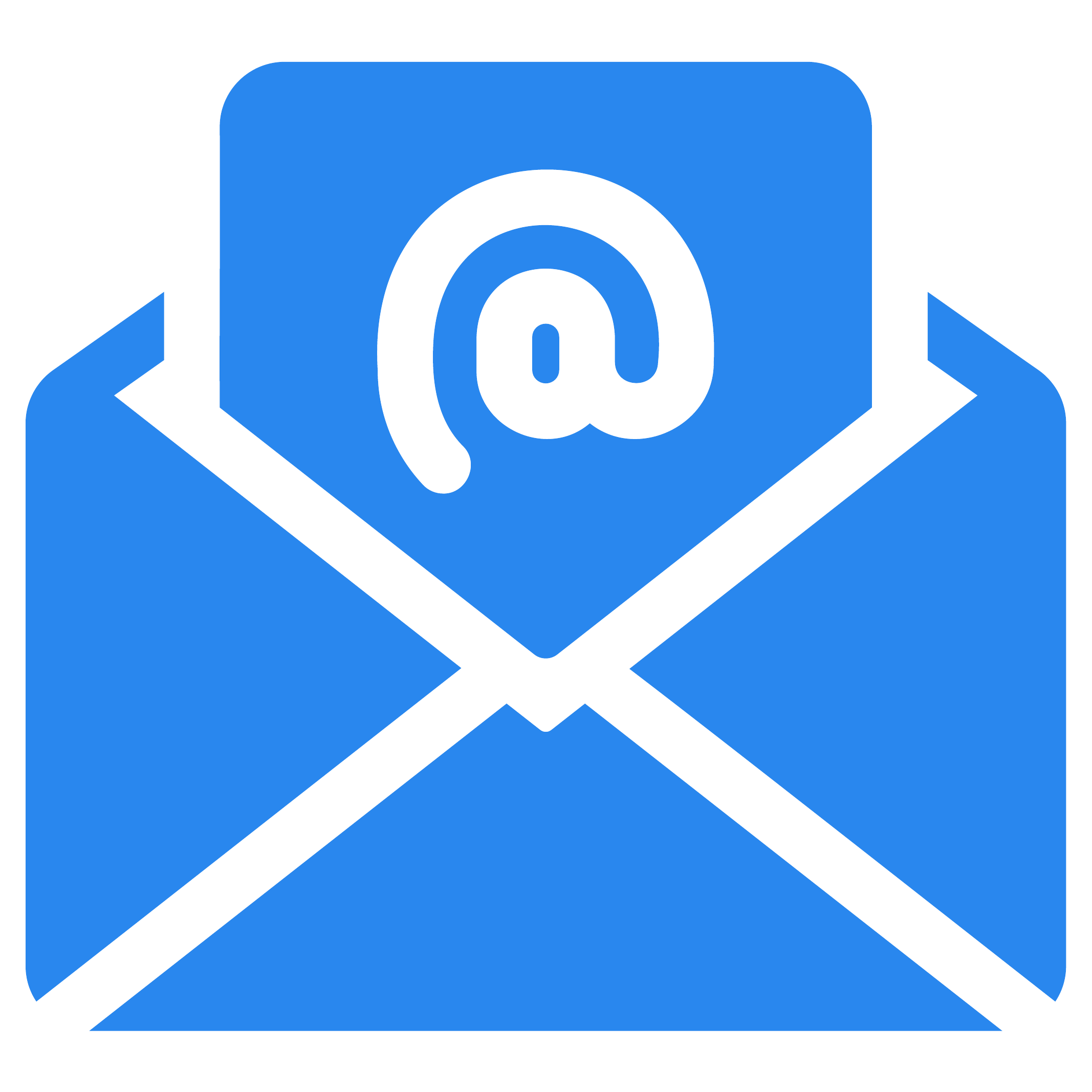}}\xspace}
\newcommand{\github}{\raisebox{-1.5pt}{\includegraphics[height=1.05em]{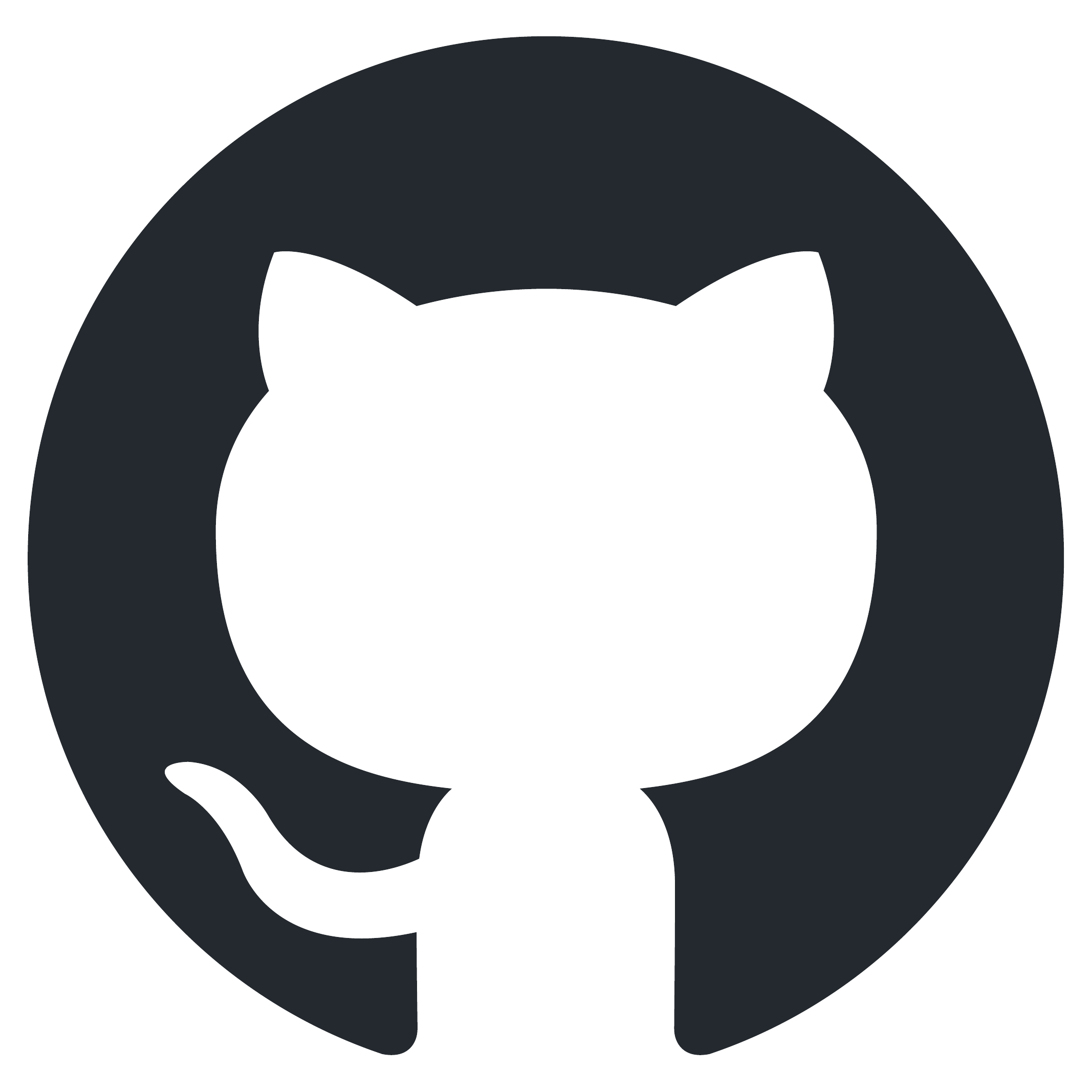}}\xspace}

\usepackage{etoolbox}
\newtoggle{anon}
\togglefalse{anon}

\def\model{OlmoEarth}
\def\platform{OlmoEarth Platform}
\def\modelurl{\url{https://github.com/allenai/olmoearth_pretrain}}

\def\datamap{figures/datamap.png}

\def\colordefault{0a3235}
\def\colorone{f0529c}
\def\colortwo{b11be8}
\def\colorthree{105257}
\def\colorfour{0fcb8c}
\def\colorfive{12cce5}
\def\colorsix{f65834}

\usepackage{multirow}
\usepackage{comment}
\usepackage{adjustbox}

\usepackage{tabularx}
\usepackage{graphicx}
\usepackage{tikz}
\usetikzlibrary{patterns.meta,positioning, arrows.meta,fit,calc,shapes.geometric,matrix}

\usepackage{pgfplots}
\pgfplotsset{compat=1.18}

\newcommand{\jm}[1]{{\color{cyan} [jacob]: #1}}

\newcommand*{\hdr}[1]{\multicolumn{1}{l}{\rlap{\,\,\,\rotatebox[origin = lb]{45}{\textbf{\scriptsize #1}}}}}

\usepackage{pdfrender}

\usepackage{graphicx}
\makeatletter
\def\maxwidth#1{\ifdim\Gin@nat@width>#1 #1\else\Gin@nat@width\fi}
\makeatother

\usepackage{contour}
\contourlength{1pt} 
\contournumber{20}  

\usepackage{pifont}

\def\ccheck{\color{black}{\ding{52}}}
\def\ccross{\color{black}{\ding{56}}}

\title{\model: Stable Latent Image Modeling for Multimodal Earth Observation}

\authorOne[\starOlmo]{Team OlmoEarth}

\authorTwo[\allenAiAff]{Henry Herzog\coreContrib}
\authorTwo[\allenAiAff]{Favyen Bastani\coreContrib}
\authorTwo[\allenAiAff]{Yawen Zhang\coreContrib}
\authorTwo[\allenAiAff]{Gabriel Tseng\coreContrib}
\authorTwo[\allenAiAff]{Joseph Redmon\coreContrib}

\authorThree[\allenAiAff]{Hadrien Sablon}
\authorThree[\allenAiAff]{Ryan Park}
\authorThree[\allenAiAff\commaAff\uwAff]{Jacob Morrison}
\authorThree[\allenAiAff]{Alexandra Buraczynski}
\authorThree[\allenAiAff]{Karen Farley}
\authorThree[\allenAiAff]{Joshua Hansen}
\authorThree[\allenAiAff]{Andrew Howe}
\authorThree[\allenAiAff]{Patrick Alan Johnson}
\authorThree[\allenAiAff]{Mark Otterlee}
\authorThree[\allenAiAff]{Ted Schmitt}
\authorThree[\allenAiAff]{Hunter Pitelka}
\authorThree[\allenAiAff]{Stephen Daspit}
\authorThree[\allenAiAff]{Rachel Ratner}
\authorThree[\allenAiAff]{Christopher Wilhelm}
\authorThree[\allenAiAff]{Sebastian Wood}
\authorThree[\allenAiAff]{Mike Jacobi}

\authorFour[\asuAff]{Hannah Kerner}
\authorFour[\ubcAff]{Evan Shelhamer}

\authorFive[\allenAiAff\commaAff\uwAff]{Ali Farhadi}
\authorFive[\allenAiAff\commaAff\uwAff]{Ranjay Krishna}
\authorFive[\allenAiAff]{Patrick Beukema\coreContrib}

\affiliation[\allenAiAff]{Allen Institute for AI}
\affiliation[\uwAff]{University of Washington}
\affiliation[\asuAff]{Arizona State University}
\affiliation[\ubcAff]{University of British Columbia}

\contribution[]{\starOlmo {\model} was a team effort.\\
\coreContrib{Equal contribution from modeling team}. Authors are listed in reverse alphabetical order by last letter of first name.}

\abstract{Earth observation data presents a unique challenge: it is spatial like images, sequential like video or text, and highly multimodal.
We present {\model}: a spatio-temporal, multimodal foundation model that employs a novel self-supervised learning formulation, masking strategy, and loss all designed for the Earth observation domain.
{\model} achieves state-of-the-art performance compared to 12 other foundation models across a variety of research benchmarks and real-world tasks from external partners. When evaluating embeddings {\model} achieves the best performance on 15 out of 24 tasks, and with full fine-tuning it is the best on 19 of 29 tasks. We deploy {\model} as the backbone of an end-to-end platform for data collection, labeling, training, and inference of Earth observation models. The {\platform} puts frontier foundation models and powerful data management tools into the hands of non-profits and NGOs working to solve the world's biggest problems. {\model} source code, training data, and pre-trained weights are available at {\modelurl}.

}

\metadata[\quad\aitoo Platform:]{\href{https://olmoearth.allenai.org/}{\texttt{olmoearth.allenai.org}}}

\metadata[\vspace{.5em}\quad\github Training Code:]{
    \href{https://github.com/allenai/olmoearth_pretrain}{\texttt{olmoearth\_pretrain}}~\sans{(pretraining)} \quad
    \href{https://github.com/allenai/olmoearth_projects}{\texttt{olmoearth\_projects}}~\sans{(fine-tuning)}
}

\metadata[\vspace{.5em}\quad\huggingface OlmoEarth Pre-trained Models:]{
    \href{https://huggingface.co/allenai/OlmoEarth-v1-Nano}{\texttt{OlmoEarth-v1-Nano}} \quad
    \href{https://huggingface.co/allenai/OlmoEarth-v1-Tiny}{\texttt{OlmoEarth-v1-Tiny}}

    \hspace{17.4em}
    \href{https://huggingface.co/allenai/OlmoEarth-v1-Base}{\texttt{OlmoEarth-v1-Base}} \quad
    \href{https://huggingface.co/allenai/OlmoEarth-v1-Large}{\texttt{OlmoEarth-v1-Large}}
}

\metadata[\vspace{.5em}\quad\huggingface OlmoEarth Pre-training Dataset:]{
\href{https://huggingface.co/datasets/allenai/olmoearth_pretrain_dataset}{\texttt{olmoearth\_pretrain\_dataset}}}

\metadata[\vspace{.5em}\quad\emailLogo Contact:]{\color{ai2accent}{\texttt{olmoearth@allenai.org}}}

\begin{document}

\definecolor{colordefault}{HTML}{\colordefault}
\definecolor{colorone}{HTML}{\colorone}
\definecolor{colortwo}{HTML}{\colortwo}
\definecolor{colorthree}{HTML}{\colorthree}
\definecolor{colorfour}{HTML}{\colorfour}
\definecolor{colorfive}{HTML}{\colorfive}
\definecolor{colorsix}{HTML}{\colorsix}
\maketitle
\newpage
\section{Introduction}
\label{sec:intro}

Earth observation foundation models show promising results in research settings \cite{astruc2024anysat, wang2025towards, fuller2024croma, tseng2025galileo, waldmann_shah_2025_panopticon, jakubik2025terramindlargescalegenerativemultimodality, feng2025tessera}.  However, adoption for real-world tasks lags behind, especially in the non-profit sector. Foundation models are large, complex to train, and expensive to deploy.

To help enable non-profit, humanitarian, and environmental organizations to use these powerful tools we train {\model}, a new family of models, using a novel, stable training regime. We comprehensively evaluate {\model} against 12 other foundation models on research benchmarks and real-world tasks from partner organizations. Finally we deploy these models in an open, end-to-end platform bringing frontier models directly to organizations who need it the most.

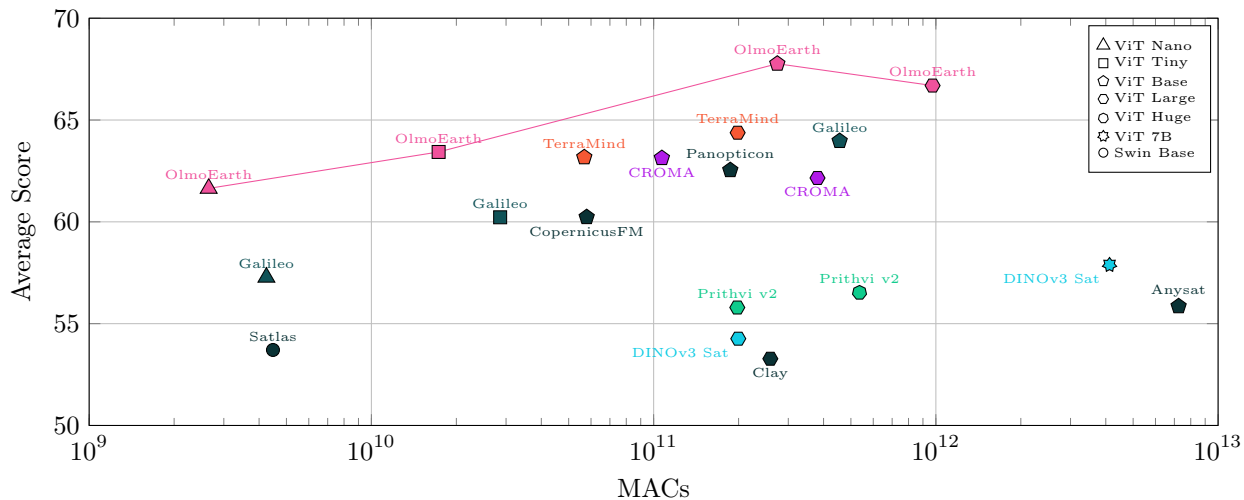
\begin{figure}[t]
    \vspace{-5pt}
    \centering
    \begin{tikzpicture}
        \begin{axis}[
            xmode=log,
            width=\columnwidth,
            height=7cm,
            xmin=1e9, xmax=1e13, 
            ymin=50, ymax=70,
            xtick={1e9, 1e10, 1e11, 1e12, 1e13},
            xlabel={MACs},
            ylabel={Average Score},
            grid=major,
        ]
        \draw[colorone,-] (2652044446, 61.64307692) -- (17309093849, 63.43153846) -- (274305011239, 67.76461538) -- (973666202545, 66.69230769);

    \matrix [
        draw,
        fill=white,
        row sep=-1mm,
        column sep=-.5mm,
        anchor=north east,
        outer sep=1mm,
        inner sep=1mm,
        column 2/.style={anchor=west},
        column 1/.style={anchor=center},
    ] at (1e13,70) {
        \node(nano)[draw, regular polygon, regular polygon sides=3, inner sep=1pt] {}; \pgfmatrixnextcell \node[font=\tiny] {ViT Nano}; \\
        \node(tiny)[draw, regular polygon, regular polygon sides=4, inner sep=1.2pt] {}; 
        \pgfmatrixnextcell
        \node[font=\tiny] {ViT Tiny};\\
        \node(base)[draw, regular polygon, regular polygon sides=5, inner sep=1.2pt] {}; 
        \pgfmatrixnextcell
        \node[font=\tiny] {ViT Base};\\
        \node(large)[draw, regular polygon, regular polygon sides=6, inner sep=1.2pt] {}; 
        \pgfmatrixnextcell
        \node[font=\tiny] {ViT Large};\\
        \node(huge)[draw, regular polygon, regular polygon sides=7, inner sep=1.2pt] {}; 
        \pgfmatrixnextcell
        \node[font=\tiny] {ViT Huge};\\
        \node(seven)[draw, star, star points=7, inner sep=1pt] {}; 
        \pgfmatrixnextcell
        \node[font=\tiny] {ViT 7B};\\
        \node(swin)[draw, circle, inner sep=1.2pt] {}; 
        \pgfmatrixnextcell
        \node[font=\tiny] {Swin Base};\\
    };

\node[regular polygon, regular polygon sides=5, inner sep=1.73329pt, draw=black, fill=colordefault] at (axis cs:7245781674318, 55.85384615) {};
\node[anchor=south, text=colordefault] at (axis cs:7245781674318, 55.85384615) {\tiny Anysat };
\node[regular polygon, regular polygon sides=6, inner sep=1.73329pt, draw=black, fill=colordefault] at (axis cs:259163629332, 53.27692308) {};
\node[anchor=north, text=colordefault] at (axis cs:259163629332, 53.27692308) {\tiny Clay };
\node[regular polygon, regular polygon sides=5, inner sep=1.73329pt, draw=black, fill=colordefault] at (axis cs:57914542789, 60.23076923) {};
\node[anchor=north, text=colordefault] at (axis cs:57914542789, 60.23076923) {\tiny CopernicusFM };
\node[regular polygon, regular polygon sides=5, inner sep=1.73329pt, draw=black, fill=colortwo] at (axis cs:106959276347, 63.13076923) {};
\node[anchor=north, text=colortwo] at (axis cs:106959276347, 63.13076923) {\tiny CROMA };
\node[regular polygon, regular polygon sides=6, inner sep=1.73329pt, draw=black, fill=colortwo] at (axis cs:380812203559, 62.15384615) {};
\node[anchor=north, text=colortwo] at (axis cs:380812203559, 62.15384615) {\tiny CROMA };
\node[regular polygon, regular polygon sides=6, inner sep=1.73329pt, draw=black, fill=colorfive] at (axis cs:199512483367, 54.26153846) {};
\node[anchor=north east, text=colorfive] at (axis cs:199512483367, 54.26153846) {\tiny DINOv3 Sat };
\node[star, star points=7, inner sep=1.3pt, draw=black, fill=colorfive] at (axis cs:4124370911232, 57.89230769) {};
\node[anchor=north east, text=colorfive] at (axis cs:4124370911232, 57.89230769) {\tiny DINOv3 Sat };
\node[regular polygon, regular polygon sides=3, inner sep=1.3pt, draw=black, fill=colorthree] at (axis cs:4246571205, 57.27692308) {};
\node[anchor=south, text=colorthree] at (axis cs:4246571205, 57.27692308) {\tiny Galileo };
\node[regular polygon, regular polygon sides=4, inner sep=1.73329pt, draw=black, fill=colorthree] at (axis cs:28588214961, 60.22307692) {};
\node[anchor=south, text=colorthree] at (axis cs:28588214961, 60.22307692) {\tiny Galileo };
\node[regular polygon, regular polygon sides=5, inner sep=1.73329pt, draw=black, fill=colorthree] at (axis cs:456133182306, 63.96923077) {};
\node[anchor=south, text=colorthree] at (axis cs:456133182306, 63.96923077) {\tiny Galileo };
\node[regular polygon, regular polygon sides=5, inner sep=1.73329pt, draw=black, fill=colordefault] at (axis cs:186936294558, 62.53846154) {};
\node[anchor=south, text=colordefault] at (axis cs:186936294558, 62.53846154) {\tiny Panopticon };
\node[regular polygon, regular polygon sides=6, inner sep=1.73329pt, draw=black, fill=colorfour] at (axis cs:197935368350, 55.79230769) {};
\node[anchor=south, text=colorfour] at (axis cs:197935368350, 55.79230769) {\tiny Prithvi v2 };
\node[regular polygon, regular polygon sides=7, inner sep=1.73329pt, draw=black, fill=colorfour] at (axis cs:536383159926, 56.52307692) {};
\node[anchor=south, text=colorfour] at (axis cs:536383159926, 56.52307692) {\tiny Prithvi v2 };
\node[circle, inner sep=1.73329pt, draw=black, fill=colordefault] at (axis cs:4480225674, 53.70769231) {};
\node[anchor=south, text=colordefault] at (axis cs:4480225674, 53.70769231) {\tiny Satlas };
\node[regular polygon, regular polygon sides=5, inner sep=1.73329pt, draw=black, fill=colorsix] at (axis cs:56794788155, 63.16923077) {};
\node[anchor=south, text=colorsix] at (axis cs:56794788155, 63.16923077) {\tiny TerraMind };
\node[regular polygon, regular polygon sides=6, inner sep=1.73329pt, draw=black, fill=colorsix] at (axis cs:198493947116, 64.37692308) {};
\node[anchor=south, text=colorsix] at (axis cs:198493947116, 64.37692308) {\tiny TerraMind };
\node[regular polygon, regular polygon sides=3, inner sep=1.3pt, draw=black, fill=colorone] at (axis cs:2652044446, 61.64307692) {};
\node[anchor=south, text=colorone] at (axis cs:2652044446, 61.64307692) {\tiny {\model} };
\node[regular polygon, regular polygon sides=4, inner sep=1.73329pt, draw=black, fill=colorone] at (axis cs:17309093849, 63.43153846) {};
\node[anchor=south, text=colorone] at (axis cs:17309093849, 63.43153846) {\tiny {\model} };
\node[regular polygon, regular polygon sides=5, inner sep=1.73329pt, draw=black, fill=colorone] at (axis cs:274305011239, 67.76461538) {};
\node[anchor=south, text=colorone] at (axis cs:274305011239, 67.76461538) {\tiny {\model} };
\node[regular polygon, regular polygon sides=6, inner sep=1.73329pt, draw=black, fill=colorone] at (axis cs:973666202545, 66.69230769) {};
\node[anchor=south, text=colorone] at (axis cs:973666202545, 66.69230769) {\tiny {\model} };

        \end{axis}
    \end{tikzpicture}
    \vspace{-10pt}

    \caption{{\model} defines a Pareto optimum of performance vs. computational efficiency averaged across 13 embedding tasks (measured by kNN and linear probing)\protect\footnotemark. The chart shows average multiply-accumulate operations to encode one example across all tasks (input size varies by task). See Table \ref{tab:knnlp} for full results.}
    \label{fig:result_summary}
\end{figure}

\subsection{Stable Training}

Foundation models are complex and expensive to train. When attempting to replicate existing work we frequently saw training instability, representation collapse, and models underperforming their stated potential. We introduce a stable training regime that models images in latent space but avoids instability and collapse.

\footnotetext{Average over all tasks every model can perform, specifically the Sentinel-2 versions of: m-bigearthnet, m-so2sat, m-brick-kiln, m-eurosat, BreizhCrops	CropHarvest-Togo, CropHarvest-PRC, m-cashewplant, m-SA-crop-type, PASTIS, MADOS, AWF, Nandi.}


Our approach strikes a middle ground between two common approaches in self-supervised learning. Masked autoencoders (\textbf{MAE}) predict pixel-level reconstructions of masked input while approaches like \textbf{I-JEPA} and Latent Masked Image Modeling (\textbf{Latent MIM}) predict reconstructions in feature space \cite{assran2023self, wei2024towards}. MAE tends to be stable but limited in its feature representations while latent approaches are unstable but produce better features \cite{mo2024connecting}.

We present Latent Masked Image Modeling of Linear, Invariant Token Embeddings (\textbf{Latent MIM Lite}), a simplification of Latent MIM that leads to stable training and better performance. We replace the target encoder of Latent MIM with a linear projection from image patches to token space that is randomly initialized and never updated during training. This simple modification stabilizes training but maintains the representative power of modeling in latent space. It also unifies self-supervised and supervised learning as we project both observational data and labeled maps through the frozen random projection layer into token space and calculate loss the same for both.

\subsubsection{Masking}


In image or text modeling it is sufficient to randomly mask some portion of the input and have the model reconstruct the input from context. With multimodal remote sensing data, any token in the input will have many similar tokens either in space, time, or at a different aligned modality. Random masking is too easy of a task unless you use a very high masking ratio \cite{tseng2025galileo}. We introduce a modality-aware masking strategy that combines random token masking with full modality reconstruction. This makes the task challenging without resorting to skewed masking ratios.

\subsubsection{Loss}

Like other SSL approaches in latent space we use a contrastive loss instead of a reconstruction loss. However, contrasting a reconstructed token against all other tokens in a batch, or even in the same sample, leads to many easy negatives given the redundant nature of Earth observation data. Instead we contrast tokens only with other tokens in their respective bandset (a subdivision of modality explained in \ref{subsec:data}). This focuses the model training on a more challenging but more productive objective, as shown in our experiments.

\subsection{Comprehensive Evaluation}

There is no standard evaluation suite for remote sensing models. While there are some established standard practices \cite{reed2023scale,fuller2024croma,tseng2025galileo}, they are not always followed. To get a more complete picture of the state of foundation modeling we run a comprehensive evaluation effort of {\model} compared to 12 other foundation models on 18 research benchmarks and 19 datasets from $7$ partner organizations that are using Earth observation modeling in their work.

Following standard practice we evaluate all models using simple transfer learning techniques (kNN and linear probing) as well as full, end-to-end fine-tuning. We evaluate all models using a standard training recipe and sweeping over a variety of hyperparameters. {\model} achieves the best performance in 15 of 24 tasks for the kNN/LP evaluation and 19 of 29 tasks for full fine-tuning. See Figure \ref{fig:result_summary} for a summary.

\subsection{Open Platform}

Training and fine-tuning remain out of reach for most environmental and humanitarian non-profits. Applying a foundation model to a task requires data gathering, alignment, pre-processing, labeling, fine-tuning, and running inference. We deploy {\model} as part of the {\platform} to simplify and streamline this process.

The {\platform} is an end-to-end solution for organizations who want to harness Earth observation data for the public good. Our partner organizations are already using the platform for things like mangrove conservation, ecosystem mapping, and food security. The {\platform} solves the last-mile problem of putting frontier research into the hands of people who can use it to do the most good.

\section{{\model}}
\label{sec:model}

{\model} is a Vision Transformer (ViT) based encoder-decoder style architecture. It processes a multimodal image timeseries of aligned satellite images and derived maps. A FlexiViT-style projection layer \cite{beyer2023flexivit} converts the input data from pixels to tokens with a variable patch size. Positional, temporal, and modality encodings add additional context to the tokens. During training, some portion of the input tokens are masked. The encoder transformer layers attend across space, time, and between modalities to produce embeddings for the input tokens. The decoder predicts representations for the masked input tokens.

\begin{figure}[t]
    \centering
    \includegraphics[width=0.9\linewidth]{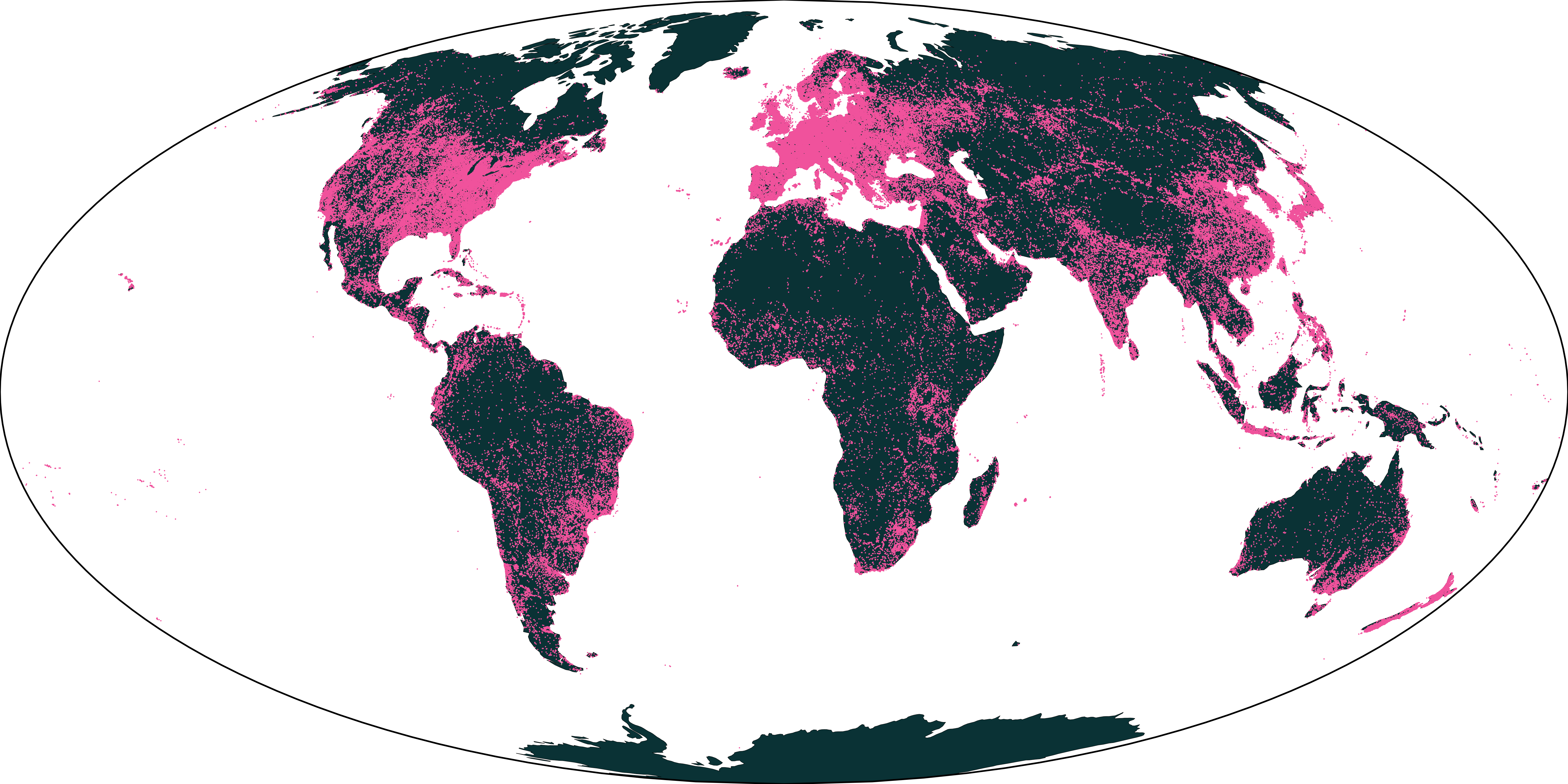}
    \caption{Global distribution of {\model} pretraining data. We sample 285,288 locations based on OpenStreetMap categories.}
    \label{fig:datamap}
\end{figure}

\subsection{Data}
\label{subsec:data}

\begin{figure*}[t]
  \centering
  \includegraphics[width=1.00\textwidth]{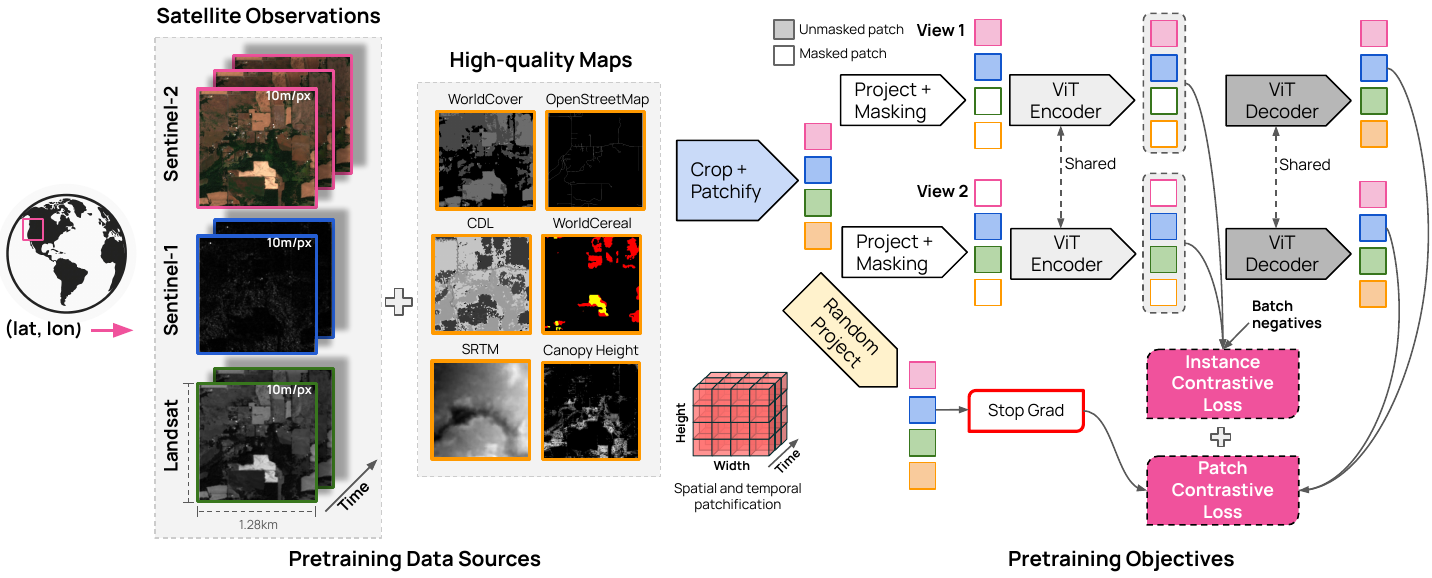}
  \caption{We train {\model} with a combination of satellite observations and high-quality maps. After tokenizing these inputs, we: (1) apply a modality-aware masking strategy to define which tokens are inputs vs. targets, (2) pass the target tokens through fixed random projections to construct targets, (3) pass the input tokens through our learned encoders, and then (4) through a decoder which predicts the target tokens and (5) apply a modality-aware patch discrimination loss between the predicted and target tokens. Steps 1-5 are applied twice on the same data to then (6) apply an instance contrastive loss over the aggregated tokens per instance.}
  \label{fig:model}
\end{figure*}

{\model} is designed to flexibly handle input Earth observation data across a range of spatial and temporal resolutions. During our pretraining experiments we train on three satellite modalities and six derived maps:

\begin{table}[h]
\centering
    \begin{tabular}{l|ll}
    Observations & Maps & \\
    \hline
    Sentinel-1 & WorldCereal \cite{van2023worldcereal} & OpenStreetMap \cite{OpenStreetMap}\\
    Sentinel-2 & WorldCover \cite{zanaga2022esa} & Cropland Data Layer \cite{USDA_NASS_Cropland_Data_Layer}\\
    Landsat-8 & SRTM \cite{SRTM} & Canopy Height Map \cite{tolan2024very}\\
    \end{tabular}
\end{table}

Our pretraining dataset contains 285,288 samples from around the world. Each sample covers a $2.56\text{km} \times 2.56\text{km}$ spatial region and a one-year time range. For multi-temporal modalities, we use up to 12 timesteps sampled monthly over the course of the year, although many samples contain only a subset of the timesteps and modalities.

For the above modalities we resample the data to be uniformly 10 meters per pixel. We experimented with adding NAIP data at 2.5 meter per pixel~\cite{NAIP_2023} and ERA5 data at 160 meters per pixel~\cite{hersbach2020era5} but found no significant improvement on our evaluations.

We further subdivide Landsat and Sentinel-2 into \textbf{bandsets} based on the original resolution of their bands, grouping bands captured at the same resolution together. Landsat consists of 2 bandsets while Sentinel-2 consists of 3 bandsets. For the precise split see the {\model} source code.

The locations of samples are chosen based on OpenStreetMap features. We select 120 categories of map features in OpenStreetMap, ranging from roads to geothermal power plants, and enumerate all $2.56\text{km} \times 2.56\text{km}$ tiles containing each category. We then randomly sample up to 10,000 tiles per category to derive the 285,288 samples (many categories appear in fewer than 10,000 tiles). The one-year time range of each sample is sampled uniformly between January 2016 and December 2024.


\subsection{Architecture}

Similar to many Earth observation models, {\model} is a transformer-based encoder-decoder style architecture. Inspired by Galileo, we use a flexible patch-embedding layer \cite{tseng2025galileo, beyer2023flexivit}. However, instead of doing that confusing pseudo-inverse stuff from FlexiViT we keep the actual projection weights the same size and resize the input image to mimic changing the patch size. It's probably basically equivalent.

Once the input is in token space, {\model} adds in a 2D sincos positional embedding, a sinusoidal temporal embedding, and a learnable modality embedding to each token. During training, some tokens are masked out of the input, otherwise all tokens are passed to the encoder transformer which performs full self-attention across space, time, and between modalities.

\begin{table}[ht]
    \centering
    {\begin{tabular}{lcccc}
       Architecture & Depth & Dim & Heads & Parameters \\
       \hline
        ViT Nano & 4 & 128 & 8 & 1.4M  \\
        ViT Tiny & 12 & 192 & 3 & 6.2M \\
        ViT Base & 12 & 768 & 12 & 90M  \\
        ViT Large & 24 & 1024 & 16 & 300M  \\
    \end{tabular}}
    
    \caption{ViT encoder architectures and number of parameters for the four {\model} model sizes.}
    \label{tab:modelsize}
\end{table}

We train four different encoder sizes based on standard Vision Transformer sizes, see Table \ref{tab:modelsize}. For each model size, the decoder has the same feature dimension and number of heads but only a depth of 4. We design a smaller decoder so that the encoder does the majority of the modeling.

During training the decoder represents the masked portions of the input with a learned $<\!\text{MASK}\!>$ token added to the appropriate positional, temporal, and modality embeddings. The decoder cross-attends to these tokens with the visible tokens from the encoder. It then predicts the latents for the masked tokens.

\subsection{Masking}

{\model} uses a modality-aware masking strategy. For every example the masking strategy selects some bandsets to be encoded and also some to be decoded, non-exclusively. Thus every bandset falls into one of four categories:

\begin{itemize}
    \item \textbf{Not selected}: Ignored for this example.
    \item \textbf{Encode only}: Randomly masked, input to encoder.
    \item \textbf{Decode only}: Used as target for decoder.
    \item \textbf{Encode and decode}: Randomly masked, input to encoder, masked tokens used as targets for decoder.
\end{itemize}

This masking strategy re-frames the problem slightly from reconstructing data that has been partially masked to reconstructing missing bandsets from partial views of other bandsets. When all bandsets are encoded and decoded we find the task is too easy. Masked tokens in a bandset will likely have other tokens in the same bandset that are highly correlated with them that are visible in the input, tokens nearby spatially or temporally. Training in this easier paradigm requires using very high masking ratios (i.e. masking out 90\% of the input) to get decent results. Masking some bandsets entirely makes the problem harder and allows more balanced masking ratios.

{\model} trains on both observations and maps but at inference time we only use observations. Maps can change over time--indeed downstream tasks are often detecting this kind of change--so we only rely on observations for inference. Thus during training our masking strategy never encodes map data, it only ever decodes it. While observations can fall into any of the above four categories, maps will only be ``decode only'' or ``not selected''.

\subsection{Latent MIM Lite}

During training {\model} predicts reconstructions of the masked input in latent space. We use a randomly initialized, frozen projection layer for each modality to project masked patches in the input into token space. Thus {\model} performs Latent Masked Image Modeling of Linear, Invariant Token Embeddings (Latent MIM Lite).

Randomly projecting raw input data extracts valuable features both from a theoretical and practical standpoint \cite{blum2005random, siddharth2020randpro, bingham2001random}. Thus our predictions are operating in a true latent space of our input data. However, because we use a fixed target encoder we avoid the representation collapse common in Latent MIM-style training. While it's possible this approach is too simplistic in more diverse domains like natural image processing, empirical results show a clear benefit in our domain of Earth observation data.

Latent MIM Lite allows us to unify supervised and self-supervised training under the same architecture. We project each modality, whether observations or maps, through a frozen random projection into token space. Loss is calculated the same for both types of modalities. We do not need to add on specific predictor heads for supervised data or adjust our training strategy or loss. In our ablations we see this approach gives strong results in a purely self-supervised setting and also benefits from additional supervised data.

Other models like Galileo and Terramind train on both supervised and unsupervised data however they treat supervised maps as a valid input to the model \cite{tseng2025galileo, jakubik2025terramindlargescalegenerativemultimodality}. This means their encoders must learn to model these map modalities as input and during training may use map modalities to predict observations or other map modalities. While this also unifies supervised and semi-supervised training, we theorize that our approach simplifies learning for the encoder while maintaining the benefits of training with supervised data. In our evaluations we see improved performance over these models on most tasks.

\subsubsection{Modality Patch Discrimination}

Masked image modeling in pixel space typically uses a reconstruction loss like Smooth L1. Latent MIM proposes using a contrastive loss (Patch Discrimination) instead of reconstruction loss to incentivize diversity in the latent space predictions. Patch discrimination loss frames token reconstruction as a classification task where we want the predicted token for a patch to be similar to the target token but dissimilar from other ground truth tokens for other patches. Patch discrimination uses cosine similarity to measure token similarity and cross entropy loss to contrast between positive and negative matches.

Typical patch discrimination contrasts a predicted token with all target tokens in the input. For image modeling, the target tokens from an image are encodings of different parts of the image so they are from the same distribution, making the contrastive task challenging. In {\model}, different target tokens can come from different modalities or different time steps as well as different spatial locations.

Tokens from different modalities have very different distributions so distinguishing between them is easy. Yet there are so many tokens from other modalities that a significant amount of the loss comes from these ``easy'' negatives. We find eliminating easy negatives and only contrasting tokens with targets from the same modality gives a substantial performance increase.

\subsubsection{Instance Contrastive Loss}

Patch discrimination loss operates on the local representations generated by the encoder and decoder but many tasks (like classification) require a global understanding of the input region. Some foundation models use a single $<\!\text{CLASS}\!>$ token to represent this global information. Instead we opt to pool information globally over all modalities, timesteps, and locations for an input. To generate a global representation for an input we run the {\model} encoder and average pool the output tokens.

Tokens encoded from the same modality share semantics but tokens from different modalities may look very different from each other. We want to be able to average tokens from all modalities together and get a sensible global representation of an input. Thus we use a contrastive loss on the pooled representation from the encoder to encourage tokens to exist in a common representation space and behave well when pooled.

We want both positive and negative samples for our contrastive loss so we take an approach similar to SimCLR \cite{chen2020simple} and encode two versions of the same input, contrasting these two versions as positive examples with the rest of the batch as negative examples. However, instead of using different data augmentation to generate the two samples we use different random masking.

We run random masking twice, then encode both batches with our encoder, pool the resulting tokens, and apply contrastive loss to the pooled representations. We run the decoder twice, decoding masked portions for both images and calculate the modality patch discrimination loss. A scalar multiple controls the contribution of instance contrastive loss to modality patch discrimination loss. For experiments in this paper we scale the instance contrastive loss by $0.1$.

\section{Experiments}
\label{sec:experiments}

We extensively evaluate {\model} on both standard research benchmarks and real-world downstream tasks from partner organizations. Following standard practice in remote sensing foundation models we evaluate both kNN/linear probe performance with a frozen encoder and full fine-tuning performance \cite{tseng2025galileo,fuller2024croma,reed2023scale}.

To get as comprehensive an evaluation as possible we import other top performing foundation models into our evaluation framework and evaluate them as well so they are directly comparable \cite{astruc2024anysat, clayfoundation, wang2025towards, fuller2024croma, siméoni2025dinov3, tseng2025galileo, waldmann_shah_2025_panopticon, tseng2023lightweight, szwarcman2024prithvi, bastani2023satlaspretrain, jakubik2025terramindlargescalegenerativemultimodality, feng2025tessera}. We use the same training recipes for each foundation model but sweep a variety of hyperparameters to find the best performance for each model on each task. We leave evaluations blank for models that do not support particular modalities. We also do not fine-tune some large models on partner tasks due to compute and time limitations.


\begin{table*}[htpb]
    \centering
    \resizebox{\textwidth}{!}{%

    
    \begin{tabular}{ll|cccccccccccccccccc|cccccc}

\hdr{} & \hdr{} & \hdr{m-bigearthnet} & \hdr{m-so2sat} & \hdr{m-brick-kiln} & \hdr{m-forestnet} & \hdr{m-eurosat} & \hdr{BreizhCrops} & \hdr{CropHarvest-PRC} & \hdr{CropHarvest-PRC} & \hdr{CropHarvest-PRC} & \hdr{CropHarvest-Togo} & \hdr{CropHarvest-Togo} & \hdr{CropHarvest-Togo} & \hdr{m-cashewplant} & \hdr{m-SA-crop-type} & \hdr{PASTIS} & \hdr{PASTIS} & \hdr{MADOS} & \hdr{Sen1Floods11} & \hdr{AWF} & \hdr{AWF} & \hdr{AWF} & \hdr{Nandi} & \hdr{Nandi} & \hdr{Nandi} \\
 & Modalities & S2 & S2 & S2 & L8 & S2 & S2 & S1 & S2 & S1,S2 & S1 & S2 & S1,S2 & S2 & S2 & S1 & S2 & S2 & S1 & L8 & S1 & S2 & L8 & S1 & S2 \\
 & Time series & \ccross & \ccross & \ccross & \ccross & \ccross & \ccheck & \ccheck & \ccheck & \ccheck & \ccheck & \ccheck & \ccheck & \ccross & \ccross & \ccheck & \ccheck & \ccross & \ccross & \ccheck & \ccheck & \ccheck & \ccheck & \ccheck & \ccheck \\
 & Method & kNN & kNN & kNN & kNN & kNN & LP & LP & LP & LP & LP & LP & LP & LP & LP & LP & LP & LP & LP & kNN & kNN & kNN & kNN & kNN & kNN \\
Model & Metric & $\mu \text{F1}$ & Acc. & Acc. & Acc. & Acc. & Acc. & Acc. & Acc. & Acc. & Acc. & Acc. & Acc. & mIOU & mIOU & mIOU & mIOU & mIOU & mIOU & Acc. & Acc. & Acc. & Acc. & Acc. & Acc. \\
\hline
Anysat & ViT Base & 54.5 & 36.5 & 84.5 & 34.0 & 80.4 & 62.7 & 57.0 & 74.1 & 72.3 & 69.6 & 78.4 & 74.5 & 24.6 & 27.2 & 24.2 & 41.9 & 41.3 & 77.8 & 60.0 & 60.0 & 64.0 & 47.4 & 20.5 & 55.8 \\
Clay & ViT Large & 48.8 & 38.7 & 90.8 & 40.3 & 86.3 & 57.0 & 56.7 & 66.5 & 63.4 & 78.4 & 67.0 & 67.3 & 30.8 & 23.1 & 19.9 & 22.6 & 47.5 & 78.9 & 59.5 & 61.5 & 56.5 & 52.2 & 23.4 & 57.0 \\
CopernicusFM & ViT Base & 64.6 & 50.3 & 85.9 & - & 84.7 & 65.5 & 55.1 & 72.8 & 74.4 & 77.5 & 75.8 & 70.6 & 32.2 & 28.4 & 15.9 & 32.1 & 63.9 & 77.6 & - & 59.0 & 67.0 & - & 24.4 & 59.8 \\
CROMA & ViT Base & 61.3 & 51.3 & 92.0 & - & 84.6 & 69.0 & 56.8 & 74.3 & 75.1 & 76.8 & 80.7 & 76.5 & 24.9 & 30.3 & 26.3 & 44.7 & 60.4 & 78.8 & - & 67.5 & \textbf{79.0} & - & 24.6 & 68.2 \\
CROMA & ViT Large & 59.2 & 48.2 & 91.7 & - & 85.8 & 68.2 & 56.2 & 72.7 & 71.0 & 79.4 & 76.5 & 81.0 & 27.0 & 30.4 & 25.9 & 42.7 & 66.4 & 78.8 & - & 62.5 & 71.0 & - & 26.1 & 68.2 \\
DINOv3 & ViT Base & 51.0 & 47.1 & 91.3 & 43.3 & 86.6 & 31.3 & - & 64.5 & - & - & 67.0 & - & 23.5 & 26.7 & - & 18.1 & 53.5 & - & 48.5 & - & 61.0 & 42.5 & - & 54.1 \\
DINOv3 & ViT Large & 55.8 & 45.3 & 89.9 & 46.0 & 84.0 & 31.3 & - & 66.1 & - & - & 68.3 & - & 24.5 & 26.1 & - & 17.4 & 52.4 & - & 43.5 & - & 60.5 & 39.9 & - & 49.8 \\
DINOv3 & ViT Huge+ & 57.0 & 45.7 & 88.2 & 46.9 & 86.1 & 31.3 & - & 68.7 & - & - & 68.3 & - & 25.1 & 26.7 & - & 17.4 & 48.1 & - & 47.5 & - & 54.0 & 43.5 & - & 55.4 \\
DINOv3 & ViT 7B & 60.8 & 46.6 & 91.3 & 48.0 & 85.2 & 31.3 & - & 68.7 & - & - & 70.9 & - & 34.3 & 27.9 & - & 21.1 & 52.2 & - & 47.5 & - & 57.0 & 44.7 & - & 56.7 \\
DINOv3 Sat & ViT Large & 60.2 & 44.0 & 91.4 & 44.2 & 89.2 & 31.3 & - & 70.1 & - & - & 68.6 & - & 32.4 & 28.5 & - & 22.5 & 57.5 & - & 42.5 & - & 69.5 & 35.5 & - & 48.4 \\
DINOv3 Sat & ViT 7B & 61.6 & 50.1 & 91.4 & 47.0 & 91.3 & 31.3 & - & 72.2 & - & - & 71.9 & - & \textbf{54.1} & \textbf{31.7} & - & 26.3 & 59.7 & - & 49.5 & - & 68.5 & 31.8 & - & 42.5 \\
Galileo & ViT Nano & 55.0 & 53.7 & 90.9 & - & 89.4 & 66.3 & 60.9 & 74.4 & 72.4 & 75.2 & 70.3 & 78.1 & 21.2 & 19.5 & 19.2 & 19.1 & 53.1 & 78.6 & - & 65.5 & 67.5 & - & 24.4 & 64.2 \\
Galileo & ViT Tiny & 55.8 & 53.1 & 87.5 & - & 89.1 & 66.7 & 55.7 & 80.3 & 79.3 & 69.9 & 78.8 & 77.1 & 23.6 & 21.5 & 23.4 & 27.7 & 61.1 & 78.6 & - & 65.5 & 71.0 & - & 25.0 & 66.7 \\
Galileo & ViT Base & 58.3 & 55.7 & 91.1 & - & 92.8 & 69.7 & \textbf{60.9} & \textbf{81.9} & 79.3 & 67.3 & 80.1 & 77.5 & 28.9 & 25.3 & 28.0 & 39.6 & \textbf{68.4} & 79.4 & - & 66.5 & 72.5 & - & \textbf{26.8} & 67.3 \\
Panopticon & ViT Base & \textbf{64.9} & 60.5 & 92.9 & \textbf{52.3} & 95.2 & 57.7 & 55.9 & 75.9 & 75.6 & 72.2 & 72.9 & 76.5 & 32.7 & 27.3 & 23.7 & 30.2 & 66.1 & 78.0 & 66.0 & 65.0 & 71.5 & 60.4 & 22.9 & 65.2 \\
Presto & ViT Nano & - & - & - & - & - & 60.9 & 59.3 & 74.3 & 76.6 & 78.4 & 81.4 & 72.5 & - & - & 16.3 & 28.2 & - & - & - & 60.5 & 53.5 & - & 25.2 & 57.6 \\
Prithvi v2 & ViT Large & 51.6 & 34.7 & 89.7 & 37.9 & 82.2 & 66.2 & - & 67.6 & - & - & 71.9 & - & 46.8 & 24.7 & - & 37.2 & 50.9 & - & 57.0 & - & 49.0 & 57.7 & - & 52.8 \\
Prithvi v2 & ViT Huge & 51.0 & 34.0 & 90.1 & 41.4 & 81.2 & 66.1 & - & 69.5 & - & - & 69.9 & - & 45.8 & 26.6 & - & 37.5 & 52.2 & - & 59.0 & - & 55.0 & 57.9 & - & 55.9 \\
Satlas & Swin Base & 52.3 & 44.5 & 83.0 & 36.9 & 82.2 & 64.6 & 57.4 & 71.3 & - & 76.8 & 75.8 & - & 30.6 & 24.0 & 10.5 & 14.4 & 30.2 & 72.9 & 57.5 & 52.0 & 62.0 & 61.5 & 25.1 & 60.2 \\
TerraMind & ViT Base & 63.9 & 46.7 & 91.9 & - & 85.6 & 66.4 & 57.0 & 75.1 & 75.6 & 74.2 & 74.8 & 77.5 & 46.0 & 30.4 & 22.7 & 40.9 & 66.0 & 78.7 & - & 66.0 & 69.5 & - & 24.7 & - \\
TerraMind & ViT Large & 63.9 & 47.4 & 92.2 & - & 90.0 & 68.2 & 56.2 & 74.7 & 72.0 & 75.2 & 77.8 & 75.2 & 50.4 & 31.2 & 22.3 & 41.3 & 67.5 & 78.4 & - & 62.0 & 67.0 & - & 23.3 & 65.3 \\
TESSERA &  & - & - & - & - & - & - & - & - & 72.2 & - & - & 81.0 & - & - & - & - & - & - & - & - & - & - & - & - \\
\hline
{\model} & ViT Nano & 59.5 & 54.3 & \textbf{96.2} & 38.8 & 89.9 & 64.1 & 58.0 & 79.5 & 74.3 & 73.5 & 81.7 & \textbf{83.7} & 25.5 & 23.6 & 18.1 & 35.0 & 55.2 & 78.2 & 73.0 & 61.5 & 69.5 & 57.7 & 24.8 & 67.4 \\
{\model} & ViT Tiny & 59.4 & 61.8 & 92.0 & 40.5 & 91.6 & 64.0 & 58.3 & 78.6 & \textbf{80.3} & 75.8 & 85.6 & 82.4 & 24.7 & 23.2 & 21.4 & 40.1 & 58.6 & 78.5 & 75.5 & 64.0 & 76.0 & 60.7 & 24.7 & 69.0 \\
{\model} & ViT Base & 62.4 & 67.7 & 93.3 & 41.9 & 94.7 & \textbf{70.9} & 56.8 & 73.4 & 75.4 & \textbf{80.1} & \textbf{87.3} & 82.0 & 32.3 & 28.9 & 29.7 & 50.6 & 67.2 & 79.2 & \textbf{77.0} & \textbf{68.5} & 77.5 & \textbf{67.9} & 26.5 & \textbf{74.7} \\
{\model} & VIT Large & 62.0 & \textbf{68.2} & 93.4 & 41.6 & \textbf{96.3} & 70.7 & 56.6 & 74.1 & 76.1 & 67.6 & 78.1 & 79.7 & 30.9 & 28.5 & \textbf{30.6} & \textbf{51.8} & 66.4 & \textbf{79.8} & 76.0 & 66.5 & 73.0 & 66.4 & 26.2 & 73.6 \\

\end{tabular}}
    \caption{kNN/Linear probe results on research benchmarks and real-world tasks from our partners. We run kNN on single time-step classification tasks and linear probing on all other tasks. We sweep across data normalization strategies, feature pooling, and learning rate (for linear probing) and report the test set result for the best validation set performance. Not all models can run on all tasks due to incompatible input modalities. {\model} has consistently strong performance and is the best on 15 out of 24 tasks.}
    \label{tab:knnlp}
\end{table*}    
\begin{table*}[htpb]
    \centering
    \resizebox{\textwidth}{!}{%
    \begin{tabular}{ll|cccccccccc|ccccccccccccccccccccc}


\hdr{} & \hdr{} & \hdr{m-bigearthnet} & \hdr{m-so2sat} & \hdr{m-brick-kiln} & \hdr{m-forestnet} & \hdr{m-eurosat} & \hdr{m-cashewplant} & \hdr{m-SA-crop-type} & \hdr{PASTIS} & \hdr{MADOS} & \hdr{Sen1Floods11} & \hdr{AWF} & \hdr{AWF} & \hdr{GEA North Africa} & \hdr{Forest Loss Driver} & \hdr{Live Fuel Moisture Content} & \hdr{Live Fuel Moisture Content} & \hdr{Mangrove} & \hdr{Mangrove} & \hdr{Marine Infrastructure} & \hdr{Marine Infrastructure} & \hdr{Nandi} & \hdr{Nandi} & \hdr{Vessel Detection} & \hdr{Vessel Detection} & \hdr{Vessel Detection} & \hdr{Vessel Length} & \hdr{Vessel Type} & \hdr{Solar Farm Detection} & \hdr{Solar Farm Detection} \\
 & Modalities & S2 & S2 & S2 & L8 & S2 & S2 & S2 & S2 & S2 & S1 & S2 & S2, S1 & S2 & S2 & S2 & S2, S1 & S2 & S2, S1 & S2 & S2, S1 & S2 & S2, S1 & L8 & S1 & S2 & S2 & S2 & S2 & S1, S2 \\
 & Time series & \ccross & \ccross & \ccross & \ccross & \ccross & \ccross & \ccross & \ccheck & \ccross & \ccross & \ccheck & \ccheck & \ccheck & \ccheck & \ccheck & \ccheck & \ccheck & \ccheck & \ccheck & \ccheck & \ccheck & \ccheck & \ccross & \ccross & \ccross & \ccross & \ccross & \ccheck & \ccheck \\
Model & Metric & $\mu \text{F1}$ & Acc. & Acc. & Acc. & Acc. & mIOU & mIOU & mIOU & mIOU & mIOU & Acc. & Acc. & Acc. & Acc. & L1 & L1 & Acc. & Acc. & F1 & F1 & Acc. & Acc. & F1 & F1 & F1 & L1 & Acc. & mIoU & mIoU \\
\hline
Anysat & ViT Base & 68.4 & 56.7 & 98.7 & 51.6 & 95.9 & 80.4 & 34.2 & 60.9 & 63.3 & 77.4 & 78.0 & 83.0 & 59.3 & 84.6 & 19.1 & 19.4 & 96.9 & 97.1 & 52.7 & 5.2 & 78.4 & 76.7 & - & - & - & 73.4 & 43.5 & 82.6 & 79.7 \\
Clay & ViT Large & 65.7 & 61.3 & 98.7 & 49.2 & 95.8 & 73.9 & 33.4 & 48.9 & 68.9 & 78.5 & 77.0 & - & 53.4 & 93.3 & 24.8 & - & 96.6 & - & 86.0 & - & 30.9 & - & 70.7 & \textbf{79.9} & 72.4 & 16.4 & 68.3 & 82.2 & - \\
CopernicusFM & ViT Base & 71.3 & 66.8 & 98.1 & - & 98.5 & 78.7 & 33.6 & 54.6 & 66.0 & 78.6 & 79.0 & 79.0 & 58.8 & 90.0 & 25.2 & 24.5 & 97.1 & 97.1 & 82.8 & 88.9 & 68.2 & 55.9 & - & 77.4 & 76.9 & 16.7 & 69.6 & 77.6 & 76.8 \\
CROMA & ViT Base & 69.5 & 59.1 & 98.7 & - & 95.6 & 46.4 & 34.8 & 56.3 & 66.6 & 79.4 & 76.5 & 75.5 & 57.5 & 93.2 & 24.3 & 24.0 & 96.4 & 96.3 & 86.1 & 84.3 & 62.2 & 67.4 & - & - & 70.0 & 19.8 & 64.4 & 79.5 & 79.5 \\
CROMA & ViT Large & 71.8 & 58.9 & 97.8 & - & 97.5 & 47.8 & 36.0 & 58.1 & 68.8 & 79.4 & 79.5 & - & 56.6 & 92.3 & 24.6 & 24.1 & 96.6 & 96.2 & - & - & 76.4 & - & - & - & - & - & - & 81.7 & - \\
DINOv3 Sat & ViT Large & 69.9 & 63.3 & 98.9 & \textbf{59.2} & 96.7 & 80.6 & 34.5 & 42.8 & 64.7 & - & 34.5 & - & 43.0 & 80.4 & 90.2 & - & 65.8 & - & 82.9 & - & 35.8 & - & - & - & 58.0 & 30.3 & 54.8 & 70.7 & - \\
Galileo & ViT Base & 69.2 & 64.7 & 98.3 & - & 97.8 & 78.8 & 35.7 & 61.2 & 71.9 & 79.7 & 81.0 & 81.5 & \textbf{62.9} & 95.1 & 20.1 & 18.7 & 97.3 & 97.5 & 85.5 & 88.0 & \textbf{81.9} & 81.9 & - & 78.7 & 75.4 & 16.4 & 73.0 & 83.1 & 85.1 \\
Panopticon & ViT Base & 69.3 & 65.4 & \textbf{99.0} & 56.0 & 98.2 & 79.7 & 33.4 & 54.4 & 72.8 & 79.1 & 75.5 & 78.5 & 54.3 & 96.4 & 24.5 & 23.7 & 97.1 & 97.4 & 86.4 & 88.6 & 65.2 & 69.5 & 74.9 & 76.7 & 76.8 & 17.7 & 69.4 & 81.8 & 79.3 \\
Prithvi v2 & ViT Huge & 70.6 & 64.7 & 98.2 & - & 96.8 & 81.1 & 38.8 & 58.6 & 69.3 & - & 80.0 & - & 60.6 & 92.4 & - & - & 97.2 & - & 84.8 & - & 77.1 & - & 71.1 & - & 74.8 & 17.4 & 68.2 & 84.1 & - \\
Satlas & Swin Base & 72.7 & 65.1 & 98.7 & 56.0 & 97.0 & 77.0 & 37.8 & 57.4 & 60.5 & 78.5 & 78.0 & - & 56.1 & 63.3 & 25.0 & 24.6 & 96.6 & - & \textbf{87.5} & - & 47.6 & - & - & - & 77.6 & 16.2 & 71.6 & 83.3 & - \\
TerraMind & ViT Base & 72.6 & 66.1 & 98.5 & - & 97.6 & 80.9 & 39.2 & 59.9 & 73.2 & 79.5 & 84.0 & 82.0 & 49.8 & 96.4 & 24.3 & 23.8 & \textbf{97.7} & 96.8 & 84.0 & 87.8 & 66.1 & 79.3 & - & 79.6 & 74.7 & 18.1 & - & 83.5 & 82.1 \\
TerraMind & ViT Large & \textbf{74.0} & 65.4 & 98.1 & - & 97.8 & \textbf{81.3} & \textbf{41.1} & 60.9 & 71.5 & 79.5 & 81.5 & - & 51.1 & 93.9 & 24.5 & 24.5 & 96.5 & 96.9 & - & - & 66.1 & - & - & - & - & - & - & 83.0 & - \\
\hline
{\model} (Random Init) & ViT Base & 61.0 & 48.9 & 94.7 & 41.7 & 80.3 & 43.0 & 27.5 & 43.9 & 45.6 & 77.0 & 62.5 & - & 52.9 & 52.7 & 20.9 & 20.5 & 96.3 & 96.4 & - & - & 60.6 & 56.1 & - & - & - & - & - & 74.1 & 70.3 \\
{\model} & ViT Nano & 66.8 & 61.5 & 98.0 & 50.3 & 95.3 & 39.5 & 35.4 & 53.0 & 60.6 & 78.8 & 82.5 & 82.5 & 61.1 & 96.0 & 20.4 & 19.7 & 97.4 & 97.4 & 86.8 & 87.8 & 75.6 & 74.8 & 70.2 & 75.5 & 75.0 & 17.1 & 72.0 & 82.1 & 79.6 \\
{\model} & ViT Tiny & 69.6 & 63.5 & 98.7 & 53.2 & 97.1 & 72.5 & 38.5 & 60.3 & 71.5 & 79.7 & 85.0 & 85.5 & 60.6 & 97.7 & 19.8 & 19.2 & 97.6 & 97.7 & 85.6 & 89.3 & 78.2 & 76.4 & 74.4 & 76.9 & 77.6 & 15.8 & 73.5 & 85.2 & 85.2 \\
{\model} & ViT Base & 72.0 & \textbf{68.6} & 98.6 & 51.2 & \textbf{98.7} & 79.8 & 39.6 & 64.3 & 77.8 & 79.8 & \textbf{87.0} & \textbf{86.0} & 62.4 & 97.1 & \textbf{18.5} & \textbf{17.9} & 97.6 & \textbf{97.9} & 86.3 & \textbf{89.6} & 81.8 & \textbf{82.2} & \textbf{75.4} & 79.2 & \textbf{78.8} & \textbf{15.4} & \textbf{74.6} & \textbf{85.4} & \textbf{86.7} \\
{\model} & ViT Large & 72.4 & 68.1 & 98.6 & 52.7 & 98.5 & 80.6 & 40.8 & \textbf{66.3} & \textbf{81.8} & \textbf{79.8} & 84.5 & 
- & 58.8 & \textbf{97.9} & 19.9 & 18.5 & 97.6 & 97.6 & - & - & 81.0 & 
- & - & - & - & - & - & 84.2 & 
- \\

\end{tabular}}
    \caption{Fine-tuning results on research benchmarks (left) and partner tasks (right). We train all models with the same recipe and report test set results for the model checkpoint with the best validation set performance. Some models are only compatible with a subset of tasks. Due to resource constraints, we do not fine-tune large models on all tasks. {\model} is best on 19 out of 29 tasks.}
    \label{tab:finetune}
\end{table*}

\subsection{Pretraining}
\label{subsec:pretrain}

We pretrain {\model} on our pretraining dataset described in \ref{subsec:data} using Latent MIM Lite. We use AdamW optimization with a base learning rate of $1\times10^{-4}$, weight decay of $0.02$, batch size of $512$, linear learning rate warm-up of $8000$ steps, cosine annealing of learning rate by $0.1$ over a total of $667,200$ steps. Due to memory constraints we use a micro-batch size of $32$ so the pooled contrastive loss is only applied over these $32$ examples, not the full batch of $512$.

During training {\model} uses a random effective patch size in the range $\{1\dots8\}$ and takes a random square crop from the input with side length in tokens in the range $\{1\dots12\}$. Thus, along the spatial dimension the smallest input is a $1\times1$ pixel region in the input with a patch size of $1$, and the largest input is $96\times96$ pixel region in the input with a patch size of $8$. Along the temporal dimension, our model processes between $3$ and $12$ timesteps. During training our model processes around 100 billion tokens. 

\subsection{Research Benchmarks and Partner Tasks} \label{sec:tasks}

We evaluate on a variety of common research benchmarks for classification and segmentation across single and multiple sensor modalities.
Our evaluations include all seven Sentinel-2 and Landsat benchmarks from GEO-Bench \cite{lacoste2024geo}: m-bigearthnet, m-so2sat, m-brick-kiln, m-forestnet, m-eurosat, m-cashewplant, and m-SA-crop-type.
We also evaluate on the classification benchmarks BreizhCrops~\cite{russwurm2019breizhcrops} and CropHarvest~\cite{tseng2021cropharvest} and the segmentation benchmarks PASTIS~\cite{garnot2021panoptic}, MADOS~\cite{kikaki2024detecting}, and Sen1Floods11~\cite{bonafilia2020sen1floods11}. 

While developing {\model}, we partnered with several organizations who are already using or want to use remote sensing data for environmental, climate, or research tasks. These organizations provided labeled data across a variety of domains for our evaluations, offering critical insights into how models perform on real-world tasks. For example, we partnered with the African Wildlife Foundation (AWF) to map land use and land cover in southern Kenya. We pair these tasks with different combinations of Sentinel-1, Sentinel-2, and Landsat observations.



\subsection{kNN and Linear Probing}

For evaluations without fine-tuning we extract embeddings from the train, validation, and test set and apply either a kNN model for single time step classification or a linear probe model for segmentation and multi-temporal classification. For {\model} we use a patch size of 4 except we sweep patch size for applicable models on m-Cashew Plant (See discussion in Appendix). For external models we use recommended settings for patch size and resize input data to that model's pretraining size following \citep{Corley_2024_CVPR}. For models that do not support time series data we input each time step separately. We sweep pooling method for the resulting embeddings across time (mean vs max). We also sweep normalization statistics (computed during pretraining vs. on the evaluation set).

We run kNN with $k=20$ using cosine similarity, and follow standard evaluation practices \cite{tseng2025galileo, gwilliam2022beyond}. For models that output a $<\!\text{CLASS}\!>$ embedding token we use that as the embedding for the whole image, otherwise we average across resulting tokens.

We run linear probing on the output embeddings, training for 50 epochs. We sweep across a variety of learning rates for each model $\{1 \times 10^{-4}, 5 \times 10^{-4},1 \times 10^{-3},5 \times 10^{-3},1 \times 10^{-2},5 \times 10^{-2},1 \times 10^{-1},5 \times 10^{-1}\}$ and report the test results for the highest validation set performance.

\subsection{Fine-Tuning}
\label{subsec:finetuning}

For fine-tuning evaluations, for each model, we take the encoder and add a decoder that makes classification, regression, semantic segmentation, or object detection predictions. Our fine-tuning recipe freezes encoder parameters for 20\% of the epochs, only training the added decoder layers, and then unfreezes and fine-tunes the full model for the remaining epochs. We use AdamW optimization with a plateau scheduler that reduces the learning rate by a factor of $0.2$ after 2 epochs without improvement on the validation set and a 10 epoch cooldown after reduction.

For fine-tuning on research benchmarks, the decoder is a single-layer linear probe; for classification tasks, it makes a prediction using embeddings pooled over the image, and for segmentation tasks, it makes a prediction using embeddings pooled temporally (when applicable) at each spatial patch. We sweep learning rates for each model over $\{1 \times 10^{-4}, 5 \times 10^{-4},1 \times 10^{-3}\}$.

For fine-tuning on partner tasks, the decoder is:
\begin{itemize}
    \item \textbf{Classification:} 3-layer MLP.
    \item \textbf{Segmentation:} Transposed convolutional layers, or U-Net decoder for multi-scale encoders \cite{ronneberger2015u}.
    \item \textbf{Object Detection:} Faster R-CNN head, with an FPN for multi-scale encoders \cite{ren2016faster, lin2017feature}.
\end{itemize}
We use a learning rate of $10^{-4}$ for all tasks, except Nandi, for which some models exhibit unstable learning and we sweep over $\{10^{-4}, 10^{-5}\}$.

\subsection{Results}

For kNN/LP evaluations, {\model} is the best performing on 11 of 18 research benchmarks and 4 of 6 partner tasks. For fine-tuning evaluations, {\model} is the best performing on 5 of 10 research tasks and 14 of 19 partner tasks. {\model} gets consistently high performance except in a couple instances.

{\model} Large does not always outperform {\model} Base, and for embedding-based pixel time series tasks it is significantly worse. This may reflect that we explore the training recipe for the Base model more than Large. Terramind and CROMA Base models often outperform Large models on many tasks so this may reflect the challenges of scaling Earth observation models.

Other notable models include Panopticon for strong performance on embedding tasks and Terramind on fine-tuning tasks. DINOv3 shows good results for tasks that mainly require visual information but lags behind specialized models on tasks where temporal understanding is critical. Galileo shows strong performance on many benchmarks, especially agriculture-related tasks.



\subsection{Ablations} \label{sec:ablations}

{

\renewcommand*{\hdr}[1]{\textbf{#1}}

\begin{table}
    \centering
    \scriptsize

    \begin{tabular}{l|ccc}

\hdr{} & \hdr{m-so2sat} & \hdr{m-eurosat} & \hdr{PASTIS} \\
\hline
Full Latent MIM* & 32.2 & 68.4 & 7.9 \\
Latent MIM Lite & 42.2 & 87.2 & 35.2 \\
+ Modality Masking & 53.6 & 90.2 & 46.6 \\
+ Modality Patch Disc & 55.3 & 91.5 & 48.1 \\
+ Contrastive Loss & 56.8 & 92.3 & 49.0 \\
+ Maps & \textbf{62.4} & \textbf{92.9} & \textbf{50.7} \\

\end{tabular}
    \caption{Development path of the {\model} base model showing effect of adding our various contributions starting from a Latent MIM approach. *Full Latent MIM collapsed during training.}
    \label{tab:development}
\end{table}
}

We based {\model} off of Latent MIM self-supervised training and iterated on various modifications, keeping the best. Table~\ref{tab:development} shows our development process, starting from standard Latent MIM, random masking, patch discrimination loss only, and no maps data. Models in the table are trained according to training recipe in Subsection \ref{subsec:pretrain} but only for 140,000 steps. Results are shown for kNN and LP on the validation set of three benchmarks. During development we ran a subset of our evaluations in our ``in-loop evals'' but saw that improvements on a representative subset carried over to the full evaluation.

We see the Latent MIM model gets poor performance due to representation collapse. Switching to Latent MIM Lite substantially boosts performance. Further modifications show increased performance for all tasks. We conduct additional ablations in Appendix \ref{app:ablations}.

\subsection{Environmental Impact}

Following recent work on environmental impact analysis of language modeling \cite{groeneveld2024olmoacceleratingsciencelanguage,morrison2025holisticallyevaluatingenvironmentalimpact,olmo20252olmo2furious} we estimate total energy use, carbon emissions, and water consumption from training {\model} in Table~\ref{tab:environmental-impact}. Similar to other environmental impact estimates this should be viewed as a lower bound as it does not account for hardware manufacturing, transportation, etc.

\begin{table}[htpb]
    \centering
\resizebox{3.25in}{!}{%
\begin{tabular}{l | c c c c c c}
 & &  & & Energy & Carbon & Water \\
Model & Stage & Hardware & GPU Hrs & (kWh) & (tCO$_2$eq) & (kL) \\
\hline
{\model} Nano & Pretraining & H100 & 1,149 & 195 & 0.08 & 0.30 \\
{\model} Tiny & Pretraining & H100 & 1,149 & 205 & 0.08 & 0.32 \\
{\model} Base & Pretraining & H100 & 2,989 & 803 & 0.32 & 1.24 \\
{\model} Large & Pretraining & B200 & 5,240 & 1,933 & 0.77 & 2.99 \\
\hline
{\model} Nano & Fine-tuning & -- & 647 & 186 & 0.07 & 0.29 \\
{\model} Tiny & Fine-tuning & -- & 723 & 261 & 0.10 & 0.40 \\
{\model} Base & Fine-tuning & -- & 1,224 & 685 & 0.27 & 1.06 \\
{\model} Large & Fine-tuning & -- & 58 & 39 & 0.02 & 0.06 \\
\hline
\textbf{Total} & Overall & -- & 13,179 & 4,307 & 1.72 & 6.67 \\
\end{tabular}
}
\caption{Approximate environmental impact of pretraining and fine-tuning {\model}. Metrics for fine-tuning {\model} Nano, Tiny, and Base include research benchmarks and partner tasks. Metrics for fine-tuning {\model} Large only include research benchmarks.
}
\label{tab:environmental-impact}
\end{table}

We train all of our models in a single data center on NVIDIA H100 and B200 GPUs. We calculate the total GPU power required for a training run by tracking actual GPU power utilization every $\sim$25ms to calculate a weighted average of power consumption throughout training. We then multiply this by the power usage efficiency (PUE) factor for our data center, according to our provider, and then we multiply this final GPU power usage amount by either the carbon intensity of the grid or the water usage efficiency factor of the data center to calculate total carbon emissions and water consumption, respectively.

The total energy usage during training (4,307 kWh) could power the average U.S. household for 5 months. The total carbon emissions are equivalent to an economy ticket on a flight from Seattle to Portugal.

\section{Related Work}
\label{sec:related}



\textbf{Pretraining} for remote sensing models initially focused on contrastive approaches \cite{jean2019tile2vec,manas2021seasonal,ayush2021geography}. Recently masked modeling has taken over as the dominant paradigm, similar to language and vision \cite{devlin2018bert,he2022masked}. Early approaches to remote sensing pretraining directly reconstructed the masked pixel values \cite{cong2022satmae,reed2023scale,tseng2023lightweight}. Following research in natural imagery \cite{assran2023self,wei2024towards,siméoni2025dinov3}, remote sensing focuses more on reconstruction in latent space. Latent approaches work well  \cite{astruc2024anysat,tseng2025galileo,waldmann_shah_2025_panopticon} but have documented instabilities \cite{mo2024connecting,assran2023self}.


\noindent\textbf{TerraMind} avoids instability by using a frozen tokenizer during pretraining. For image modalities they train a quantized autoencoder and use the encoder as their frozen tokenizer during multimodal masked modeling.

\noindent\textbf{Precomputed embeddings} offer an alternative approach for accessibility \cite{brown2025alphaearth,feng2025tessera} but still require expertise to retrieve and use. Best results may still require training a decoder on top of the embeddings. Precomputed embeddings also limit flexibility; both AEF and TESSERA generate annualized embeddings making real-time or sub-annual predictions impossible. {\model} embeddings match or outperform AEF embeddings on partner tasks, and full fine-tuning enables even better results (Appendix Table \ref{tab:aef}).

\section{Discussion}

We want {\model} to have a positive impact on the world. Toward that end we release it as part of the {\platform}, an end-to-end, open solution for Earth observation tasks. {\platform} enables partner organizations to use the latest, best foundation models in their work on the environment, conservation, food security, and more. Organizations like Global Mangrove Watch, Global Ecosystem Atlas, and the International Food Policy Research Institute are using {\platform} for data curation and labeling, model fine-tuning, and inference.

\subsection{Case Studies}

\paragraph{Global Mangrove Watch} maps and tracks the extent and health of coastal mangrove forests. Mangrove forests sequester carbon, protect the coastline from erosion, and provide a habitat for little fishies. GMW uses a random forest model with a $95.3\%$ F1 score to generate maps on a yearly cadence, and only covering about half of relevant coastal regions. Using {\platform} we fine-tune a {\model} model using their data up to an F1 score of $98.1\%$. The {\platform} can run inference on a monthly cadence to generate new maps, or on a rolling basis to detect change faster.

\begin{figure}[ht]
  \centering
  \includegraphics[width=\columnwidth]{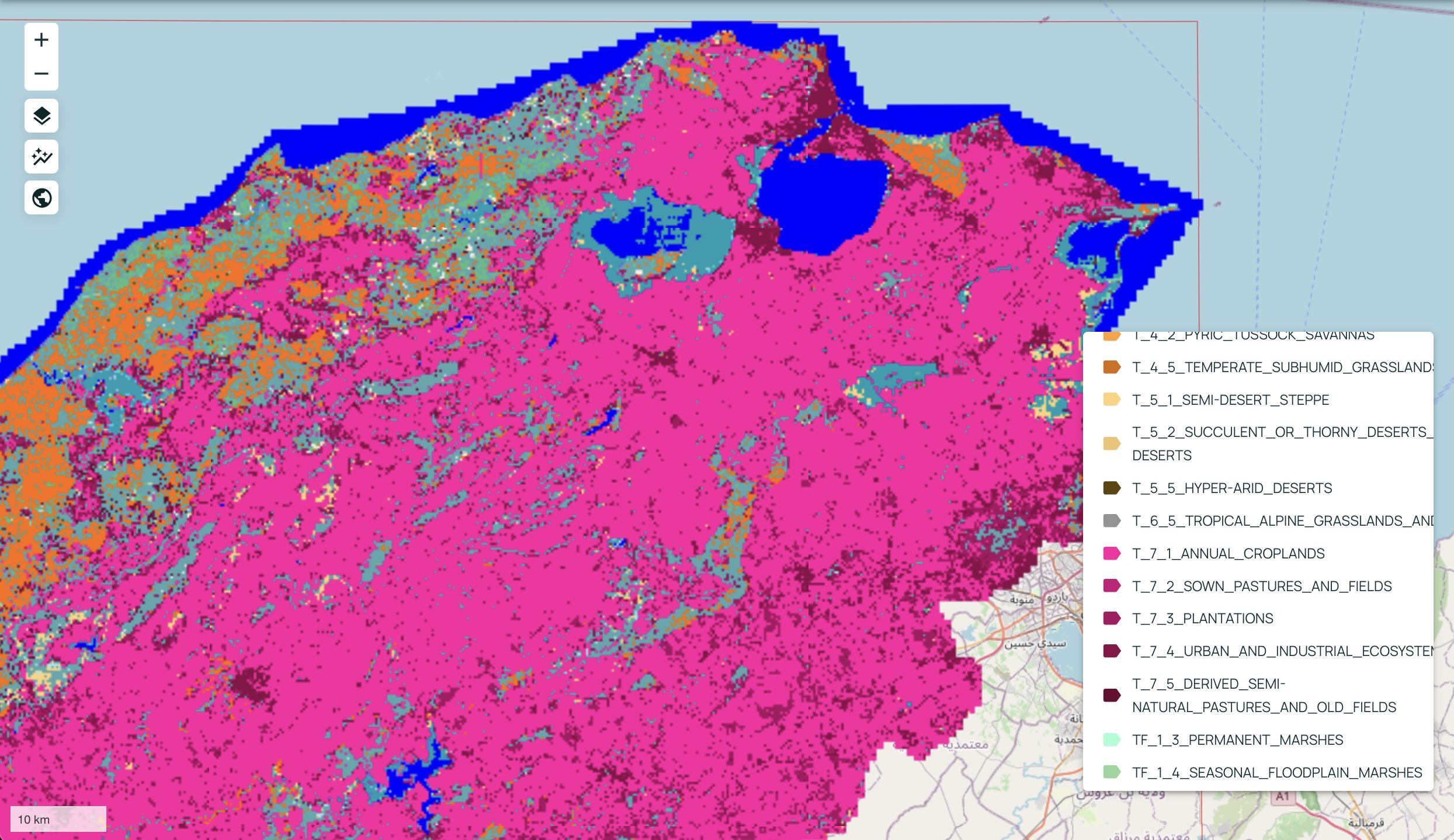}
  \caption{Results of a fine-tuned ecosystem classification model in the {\platform}. Users can label data, fine-tune models, and run inference to generate maps all in the {\platform}.}
  \label{fig:gea}
\end{figure}

\paragraph{Global Ecosystem Atlas} is building a comprehensive map of the world's ecosystems \cite{Keith2022A}. For the last 3 months they have been using {\platform} to label more than 15,000 data points. {\platform} allows them to partition areas of interest, generate points to label, assign those points to labelers, review the results, and export the data or fine-tune a model directly in the platform. With a subset of the data from North Africa we fine-tune a model that achieves state-of-the-art accuracy and run inference to generate new ecosystem maps. Humans can review the results to feed better labels back into the training pipeline.

\subsection{Downstream Risks}
The power and versatility of {\model} also bring risks.
We release {\model} under an open license designed to address some of these risks by allowing the free use, modification, and sharing of the model weights, datasets, and associated code while restricting use for military, defense-related, and extractive industry applications.


\subsection{The Future}

We plan to add climate and weather data and forecasting to the {\model} model to help with tasks like wildfire prediction and crop yield forecasting. Expanding to this kind of data will require handling a wider variety of input resolutions both spatially (from meters to kilometers) and temporally (from days to years).

We also plan to add non-geospatial data to the model. Often data labeling for tasks like crop type mapping requires actually going to a location in person and looking at stuff. We'd like the model to be able to do that too. The ability to process geolocated natural images would expand {\model}'s ability to handle these fine-grained recognition tasks.

Ultimately we want to support and grow the community of partner organizations who bring incredible knowledge, expertise, and passion to this work. We plan to learn from our partners about what tools and capabilities they need and then improve {\platform} to better help them. We hope {\platform} can become a hub for data, models, training, and inference across a wide range of organizations working to solve the world's biggest problems.

\section*{Acknowledgments}

We wish to express deep gratitude to our early collaborators who shared data, expertise, and time to make these models successful for real-world, mission-critical applications: Amazon Conservation Association, African Wildlife Foundation, CGIAR/International Food Policy Research Institute (IFPRI), Global Mangrove Watch, Global Ecosystem Atlas, ITC University of Twente, NASA Jet Propulsion Laboratory (JPL), and NASA Harvest.

We would also like to thank the OLMo-core, Beaker, Comms, and Legal teams at Ai2 for their support, especially Pete Walsh, Dirk Groeneveld, Sam Skjonsberg, Tara Wilkins, Caroline Wu, Johann Dahm, David Albright, Kyle Wiggers, Jordan Steward, Crystal Nam, Will Smith, and Janice Dow.

\clearpage
\bibliographystyle{ieeenat_fullname}
\bibliography{main}

\clearpage

\appendix

\section{Research Benchmarks} \label{app:research_benchmarks}

We describe the research benchmarks introduced in Section \ref{sec:tasks} in more detail below.
We also share our observations on limitations of certain benchmarks.

GEO-Bench modifies benchmarks to form a unified and consistent collection of datasets:

\textbf{m-bigearthnet} is modified from BigEarthNet~\cite{sumbul2019bigearthnet}, which involves multi-label land cover classification of $120 \times 120$ Sentinel-2 image crops. It consists of 19 classes, such as arable land, inland wetlands, and urban fabric. The original dataset contains 549,488 examples, but the modified subset in GEO-Bench contains only 22,000, with 20,000 for training, 1,000 for validation, and 1,000 for testing.

\textbf{m-so2sat} is modified from So2Sat LCZ42~\cite{zhu2020so2sat}, which involves image-level classification of local climate zones from co-registered Sentinel-1 and Sentinel-2 crops. It consists of 17 classes, such as high-rise, industrial, and water bodies. The original dataset contains 400,673 examples, but the modified subset in GEO-Bench contains only 21,964, with 19,992 for training, 986 for validation, and 986 for testing.

\textbf{m-brick-kiln} is modified from the Brick Kiln Classification Dataset in Bangladesh~\cite{lee2021scalable}. The original dataset involves image-level classification of whether or not high-resolution $224 \times 224$ satellite image crops from DigitalGlobe contain at least one kiln, and contains 6,329 positive examples and 67,284 negative examples. The modified dataset in GEO-Bench performs the same task on corresponding $64 \times 64$ Sentinel-2 crops, and contains only 17,061 examples, with 15,063 for training, 999 for validation, and 999 for testing. While finding kilns in Sentinel-2 images is a challenging task, we find that the nature of the negatives in the GEO-Bench version of the dataset make the classification task too easy; for example, many negatives seem to have only dark pixels, making it easy to distinguish them.

\textbf{m-forestnet} is modified from ForestNet~\cite{irvin2020forestnet}, which involves image-level classification of deforestation drivers from a composite $332 \times 332$ Landsat 8 satellite image captured within five years after each forest loss event. There are four driver categories: plantation, smallholder agriculture, grassland/shrubland, and other. The original dataset contains 2,756 examples. The modified subset in GEO-Bench contains 6,464 examples for training, 989 examples for validation, and 993 examples for testing; we could not determine where the additional examples came from.

\textbf{m-eurosat} is modified from EuroSat~\cite{helber2019eurosat}, which involves image-level land use and land cover classification from $64 \times 64$ Sentinel-2 image crops. It consists of 10 classes, such as annual crop, river, and highway. The original dataset contains 27,000 examples, but the modified subset in GEO-Bench contains only 4,000, with 2,000 for training, 1,000 for validation, and 1,000 for testing.

\textbf{m-cashewplant} is modified from the Smallholder Cashew Plantations in Benin Dataset~\cite{smallholdercashew}, which involves segmentation of $256 \times 256$ Sentinel-2 image crops. It consists of six classes relating to cashew plantations: well-managed plantation, poorly-managed plantation, non-plantation, uncertain, residential, and background. The modified dataset contains 1,800 examples, with 1,350 for training, 400 for validation, and 50 for testing. Multiple models were sensitive to input patch size on this dataset, so for models that had a variable patch size, we swept input patch size and report the best result. Ultimately this is likely an effect of the labels being large polygons instead of per-pixel labels.

\textbf{m-SA-crop-type} is modified from the South Africa Crop Type Competition dataset~\cite{sacroptype}, which involves crop type segmentation of $256 \times 256$ Sentinel-2 and Sentinel-1 image crops. It consists of 10 classes, such as fallow, wine grapes, and wheat. The modified dataset in GEO-Bench only uses the Sentinel-2 images, and contains 5,000 examples, with 3,000 for training, 1,000 for validation, and 1,000 for testing.

All of the GEO-Bench datasets share a significant limitation: although the tasks involve labels that do not change rapidly over time, the input consists of a single satellite image or image pair. We find that remote sensing models generally perform much better with multiple input images, and argue that single-image inputs should only be used for tasks like vessel detection where the labels are only valid for one timestep.

We compare on five additional datasets outside of GEO-Bench:

\textbf{BreizhCrops}~\cite{russwurm2019breizhcrops} involves crop type classification from single-pixel Sentinel-2 time series. It consists of nine classes, such as wheat, corn, and permanent meadows. It contains 610K examples.

\textbf{CropHarvest}~\cite{tseng2021cropharvest} involves binary cropland classification from single-pixel time series. The provided time series include Sentinel-2 and Sentinel-1 satellite image observations, as well as elevation from SRTM and weather data from ERA-5. It contains 95,186 examples.

\textbf{PASTIS}~\cite{garnot2021panoptic} involves crop type segmentation from Sentinel-1 and Sentinel-2 image time series, with $128 \times 128$ image crops. It consists of 19 classes, such as grapevine, spring barley, and soybeans. It contains 2,433 examples.

\textbf{MADOS}~\cite{kikaki2024detecting} involves marine debris segmentation in $80 \times 80$ Sentinel-2 image crops. It consists of 15 classes, such as oil spills, dense sargassum, and foam. It contains 2,803 examples. 
A key limitation with MADOS is that it provides custom-processed images, making it difficult to apply foundation models with their intended normalization statistics. Additionally, the dataset includes a lot of rare classes that greatly affect mIoU in the test set, making metrics highly variable across runs of the same model with different seeds.

\textbf{Sen1Floods11} involves binary water segmentation in $512 \times 512$ Sentinel-2 image crops that focus on flooded areas. It contains 4,831 examples. All of the remote sensing models we tested get between 78-80\% accuracy, and we find that the accuracy is not well correlated with other benchmarks. However, Sen1Floods11 is one of the few Sentinel-1 benchmarks.

\section{Partner Tasks} \label{app:partner_tasks}

We describe the partner tasks introduced in Section \ref{sec:tasks} in more detail below.

\textbf{AWF - \textit{African Wildlife Foundation (AWF)}} Land cover classification in southern Kenya. The dataset contains 1,459 examples with 9 classes, which range from lava forest and agriculture to urban development. The AWF team used Planet imagery as the main reference to annotate these examples.

\textbf{Live Fuel Moisture Content - \textit{NASA JPL}} Regression dataset of 41,214 examples from Globe-LFMC-2.0~\cite{globelfmc} labeled with the LFMC value. We partner with NASA JPL to deploy a model trained on this data. LFMC predictions are used to understand wildfire risk.

\textbf{Mangrove - \textit{Global Mangrove Watch}} Classification dataset of 100,000 coastal areas into 3 classes: mangrove forest, water, or other. Mangrove maps across different years are used to understand mangrove growth and loss.

\textbf{Nandi - \textit{CGIAR}} Crop-type classification in Nandi County, Kenya. The dataset contains 6,924 examples with 6 categories (coffee, maize, sugarcane, etc.). The ground-truth labels were collected through field surveys.


Ecosystem type mapping is similar, but only uses six timesteps of input images:

\textbf{GEA North Africa - \textit{Global Ecosystem Atlas}} Ecosystem type classification of 2,361 examples in a region of North Africa, and labels correspond to the 110 categories in level 3 of the IUCN Global Ecosystem Typology \cite{GEO_Atlas_2025}.

The other tasks are more unique:

\textbf{Forest Loss Driver - \textit{Amazon Conservation}} Classification dataset for the cause of forest loss in the Amazon rainforest into 10 classes (mining, logging, agriculture, etc.). The input consists of 4 Sentinel-2 images captured before the forest loss and 4 images captured after the forest loss. Driver predictions are used to prioritize enforcement and litigation efforts to deter further human-caused forest loss.

\textbf{Marine Infrastructure - \textit{Skylight}} Global marine infrastructure detection dataset containing 7,197 examples labeled as offshore platform or wind turbine. The input consists of a time series of 4 Sentinel-2 or Sentinel-2 + Sentinel-1 images.

\textbf{Vessel Detection, Type, Length - \textit{Skylight}} Three object detection tasks to detect vessels in Landsat (8,000 examples), Sentinel-1 (1,776 examples), and Sentinel-2 (45,545 examples) images, one classification task to predict the vessel type in Sentinel-2 images centered at detected vessels (584,432 examples), and one regression task to estimate the vessel length in Sentinel-2 images (584,432 examples). For all of these tasks, the input is a single image.

\textbf{Solar Farm Detection}: Binary segmentation dataset containing 3,561 examples densely labeled with solar farm polygons. The input consists of 4 timesteps, either Sentinel-2 or Sentinel-2 + Sentinel-1. Solar farm maps are used to understand the global rate of renewable energy deployment over time.

\section{Additional Ablations} \label{app:ablations}

\begin{table*}[ht]
    \centering
    \resizebox{0.75\linewidth}{!}{%
    \begin{tabular}{l|cccccccccc|cc}

\hdr{} & \hdr{m-bigearthnet} & \hdr{m-so2sat} & \hdr{m-brick-kiln} & \hdr{m-forestnet} & \hdr{m-eurosat} & \hdr{BreizhCrops} & \hdr{PASTIS} & \hdr{PASTIS} & \hdr{MADOS} & \hdr{Sen1Floods11} & \hdr{Average} & \hdr{Average Rank} \\
 & S2 & S2 & S2 & L8 & S2 & S2 & S1 & S2 & S2 & S1 &  &  \\
 & Acc. & Acc. & Acc. & Acc. & Acc. & F1 & F1 & F1 & F1 & F1 &  &  \\
\hline
MAE & 60.6 & 48.1 & 96.2 & 42.0 & 89.3 & 71.5 & \textbf{31.1} & 46.6 & 68.7 & 78.3 & 63.2 & 5.1 \\
\hline
Only S2 Data & 53.7 & 45.9 & 91.3 & - & 89.2 & \textbf{71.7} & - & 42.4 & 69.5 & - & 46.4 & - \\
No Maps & 59.5 & 58.6 & 95.2 & \textbf{46.0} & 92.6 & 71.4 & 29.1 & 48.0 & 70.2 & 77.9 & 64.9 & 4.7 \\
No Agricultural Maps & 60.9 & 66.5 & 94.3 & \textbf{46.0} & 93.9 & 71.4 & 29.0 & 48.5 & 71.4 & 78.8 & 66.1 & 3.6 \\
\hline
Random Masking & 60.7 & \textbf{67.4} & 94.7 & 43.5 & 91.8 & 70.3 & 24.7 & 51.1 & 71.9 & 77.8 & 65.4 & 4.7 \\
No Inst. Contrastive Loss & 60.5 & 65.6 & 93.6 & 44.9 & 93.6 & 70.2 & 28.5 & 51.4 & 72.1 & 78.4 & 65.9 & 4.7 \\
Patch Disc Loss & 62.0 & 62.1 & \textbf{96.3} & 44.8 & 94.0 & 70.3 & 29.6 & 50.0 & \textbf{74.1} & \textbf{79.3} & 66.2 & 3.0 \\
\hline
Final Recipe & \textbf{62.3} & 65.9 & 94.2 & 45.8 & \textbf{94.6} & 71.4 & 29.4 & \textbf{52.2} & 71.7 & 78.8 & \textbf{66.6} & \textbf{2.9} \\

\end{tabular}
}
    \caption{Ablation experiment selectively removing components of {\model} base model.}
    \label{tab:ablation}
\end{table*}

In addition to the ablations in Section \ref{sec:ablations}, we conduct a second set of ablations in Table \ref{tab:ablation}. Our second set of ablations evaluates the contributions of components of our final model and training recipe by removing them individually, with the exception of the top row which is a MAE baseline. These models are trained for 300,000 steps. In the data ablation section we see the Sentinel-2 only model perform relatively poorly, however the ``No Maps" run (only observational data) maintains relatively high performance. While our model can benefit from labeled data we still see good performance with pure self-supervised training.

Building remote sensing foundation models necessitates some tradeoffs. While our final model is not the best in every metric it retains high performance across the board and has the best average score and lowest average per-task rank.

\section{Comparison to AlphaEarth Foundations}

The AlphaEarth foundation model~\cite{brown2025alphaearth} is comparable to {\model} in that both draw on similar data sources and were designed to support similar downstream tasks. Rather than releasing the model, Google released only the global, annualized embeddings computed by AlphaEarth. We compare {\model} both as a frozen feature extractor (where, like AlphaEarth, only embeddings are used) and as an end-to-end finetune-able model.

\begin{table*}
    \setlength{\tabcolsep}{4pt}
    \centering
    \begin{tabular}{ll|ccccc}

\hdr{} & \hdr{} & \hdr{Nandi} & \hdr{AWF} & \hdr{Ecosystem} & \hdr{LFMC} & \hdr{Solar Farm} \\
Model & Training & Acc. & Acc. & Acc. & L1 & mIOU \\
\hline
AEF & kNN & 55.6 & 81 & 60.6 & - & - \\
AEF & Frozen + Decoder & 66.0 & 75.9 & 61.2 & 23.1 & 77.5 \\
AEF & Full Fine-tuning & \multicolumn{5}{c}{Not Possible} \\
\hline
{\model} & kNN & 66.2 & 82 & 59.3 & - & - \\
{\model} & Frozen + Decoder & 62.9 & 84.0 & 61.1 & 19.9 & 84.8 \\
{\model} & Full Fine-tuning & \textbf{82.2} & \textbf{86.0} & \textbf{62.4} & \textbf{17.9} & \textbf{86.7} \\

\end{tabular}
\caption{Comparing AlphaEarth Foundation (AEF) embeddings with {\model} ViT Base model using three different training strategies: kNN, frozen backbone + decoder, and decoder with full fine-tuning. For these evaluations, we use the ``partner task'' decoders described in Section \ref{subsec:finetuning}.} \label{tab:aef}
\end{table*}

It is expensive to export and download AlphaEarth embeddings from Google Earth Engine: our export jobs for $32 \times 32$ crops took 26 EECU-seconds on average, or \$290 for a dataset with 100K crops.
Thus, we were only able to evaluate AlphaEarth on five tasks: three classification tasks (Nandi, AWF, and Ecosystem), one per-pixel regression task (LFMC), and one segmentation task (Solar Farm).

Since the AlphaEarth model has not been released, we can't evaluate AlphaEarth under a finetuning regime. We assess the performance of the annualized AlphaEarth embeddings compared to the {\model} embeddings from the ViT Base encoder using a simple KNN classifier. We use the timestep of AlphaEarth embeddings that has the highest overlap with the time range of the labels. To assess the benefits of more complex decoders, we use the partner task decoders described in Section \ref{subsec:finetuning}, while sweeping over the input size (AlphaEarth embeddings already capture spatial context, so we find that a smaller input size performs better).

With a KNN-classifier, {\model} outperforms AlphaEarth on the Nandi and AWF tasks, while AEF outperforms {\model} on the Ecosystem mapping task. However, {\model} benefits significantly from full fine-tuning, with the fine-tuned models outperforming the best possible with AlphaEarth on all five tasks. This underscores the value of an open model that makes per-task fine-tuning possible.

\section{Patch Size Analysis for \texttt{m\_cashew\_plant}}

We observe that for the \texttt{m\_cashew\_plant} evaluation task, larger patch sizes lead to better performance for models that support variable patch sizes, such as {\model} and Galileo. Table~\ref{tab:patchsize_performance} summarizes the linear probing and fine-tuning results for \texttt{m\_cashew\_plant} across different patch sizes.

\begin{figure}[t]
    \centering
    \includegraphics[width=0.5\linewidth]{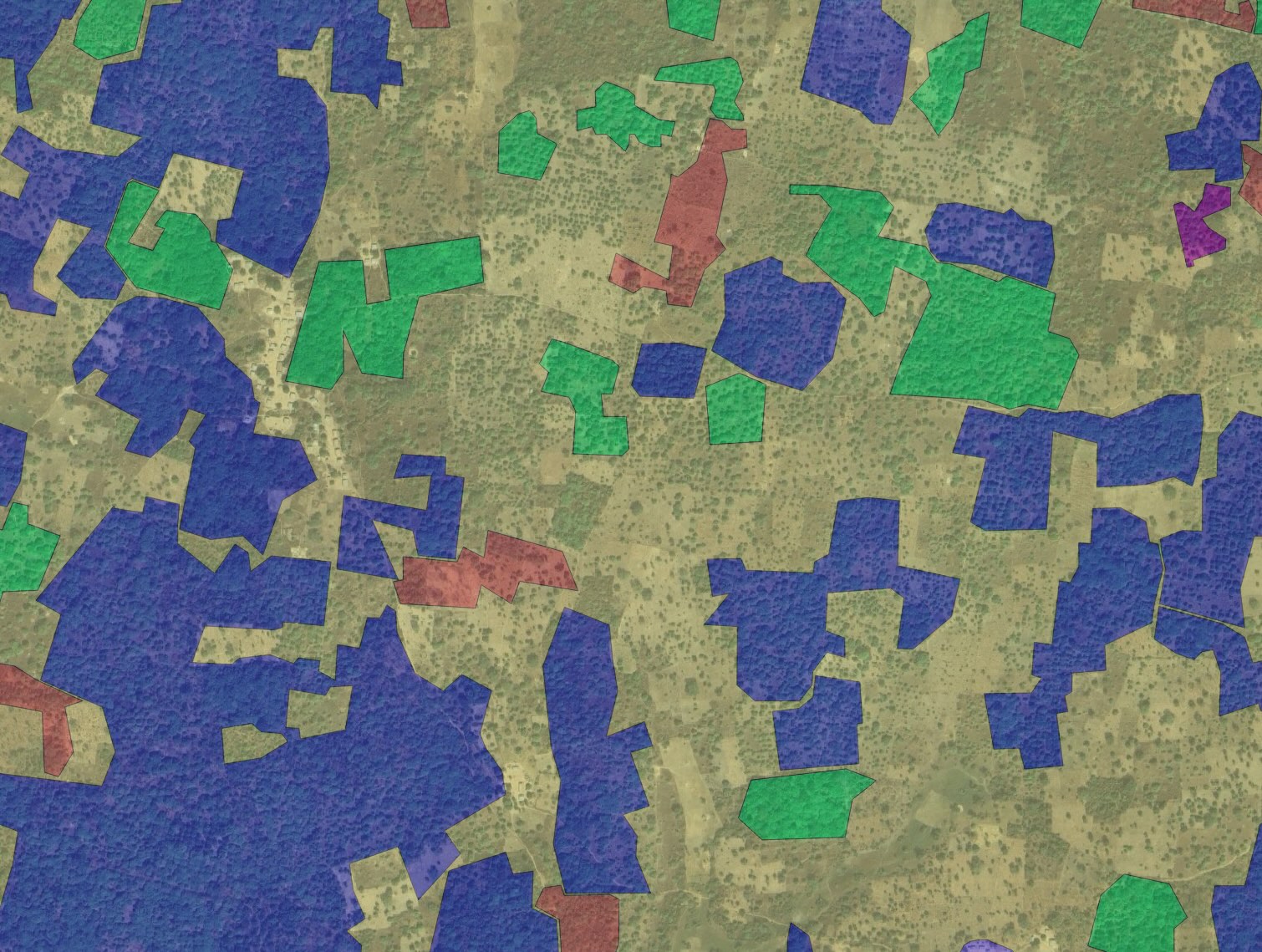}
    \caption{An example instance from the \texttt{m\_cashew\_plant} dataset: note the coarse, polygonal labels}
    \label{fig:cashew-plant}
\end{figure}

This effect is unusual: a smaller patch size typically improves performance (e.g. Figure 4 of \cite{tseng2025galileo}). We hypothesize that this is due to the spatially coarse labels in the dataset, which are polygons instead of pixels (Figure \ref{fig:cashew-plant}).

\begin{table}[ht]
\centering
\begin{tabular}{lcccccc}
\toprule
\multirow{2}{*}{Model} & \multicolumn{2}{c}{Patch 4$\times$4} & \multicolumn{2}{c}{Patch 8$\times$8} & \multicolumn{2}{c}{Patch 16$\times$16} \\
\cmidrule(lr){2-3} \cmidrule(lr){4-5} \cmidrule(lr){6-7}
 & LP & FT & LP & FT & LP & FT \\
\midrule
{\model}-Base & 27.7 & 71.9 & 27.9 & 76.2 & 32.3 & 79.8 \\
Galileo   & 24.3 & 73.0 & 25.6 & 76.9 & 28.9 & 78.8 \\
\bottomrule
\end{tabular}
\caption{Performance (mIoU) comparison (LP = Linear Probing, FT = Fine-tuning) across patch sizes.}
\label{tab:patchsize_performance}
\end{table}

\begin{figure*}[t]
    \centering
    \includegraphics[width=\textwidth]{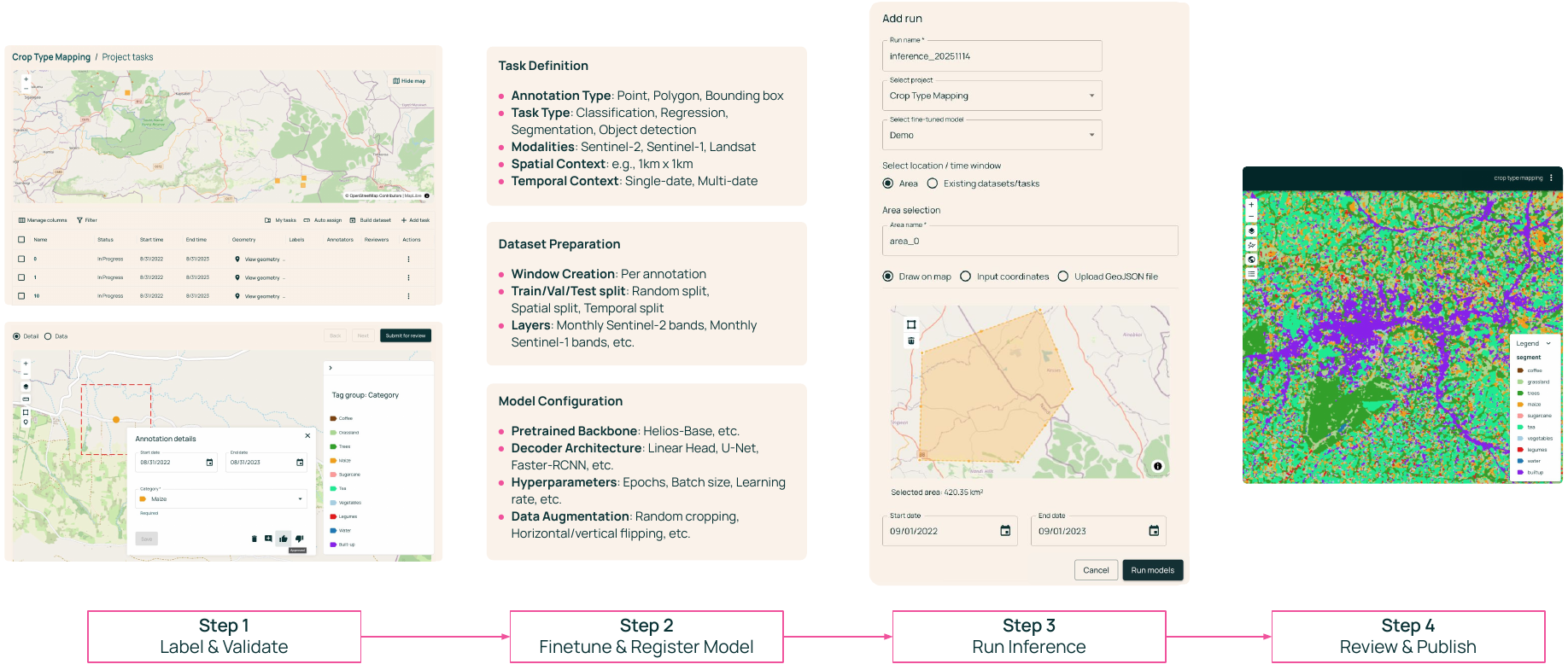}
    \caption{
        \textbf{{\platform}: End-to-End Workflow (using crop type mapping as an example).}
        The platform enables users to complete the full process from data labeling to map publishing:
        \textbf{Step 1:} Label and review annotations, 
        \textbf{Step 2:} Fine-tune and register models for specific tasks, 
        \textbf{Step 3:} Run inference on selected areas and time ranges, and 
        \textbf{Step 4:} Review and publish the final maps.
    }
    \label{fig:helios_platform}
\end{figure*}

\section{{\platform}}
{\platform} is an end-to-end solution that combines our foundation models with data management tools designed for organizations working on environmental challenges. The platform handles the complete workflow from satellite data collection through labeling, model fine-tuning, and inference, eliminating the need for organizations to manage GPU infrastructure or deep learning expertise. By making our models accessible, {\platform} solves the last-mile problem of translating research into practical tools for applications including conservation, climate action, and food security.







\iftoggle{anon}{}{
    \clearpage
    \definecolor{antcolor}{HTML}{0a3235}
    \captionsetup{font={color=antcolor, small}}
    
    \begin{figure*}[p!]
    \begin{tikzpicture}[remember picture, overlay]
      \node[inner sep=0,text width=\textwidth] at (current page.center){
        \includegraphics[width=\textwidth]{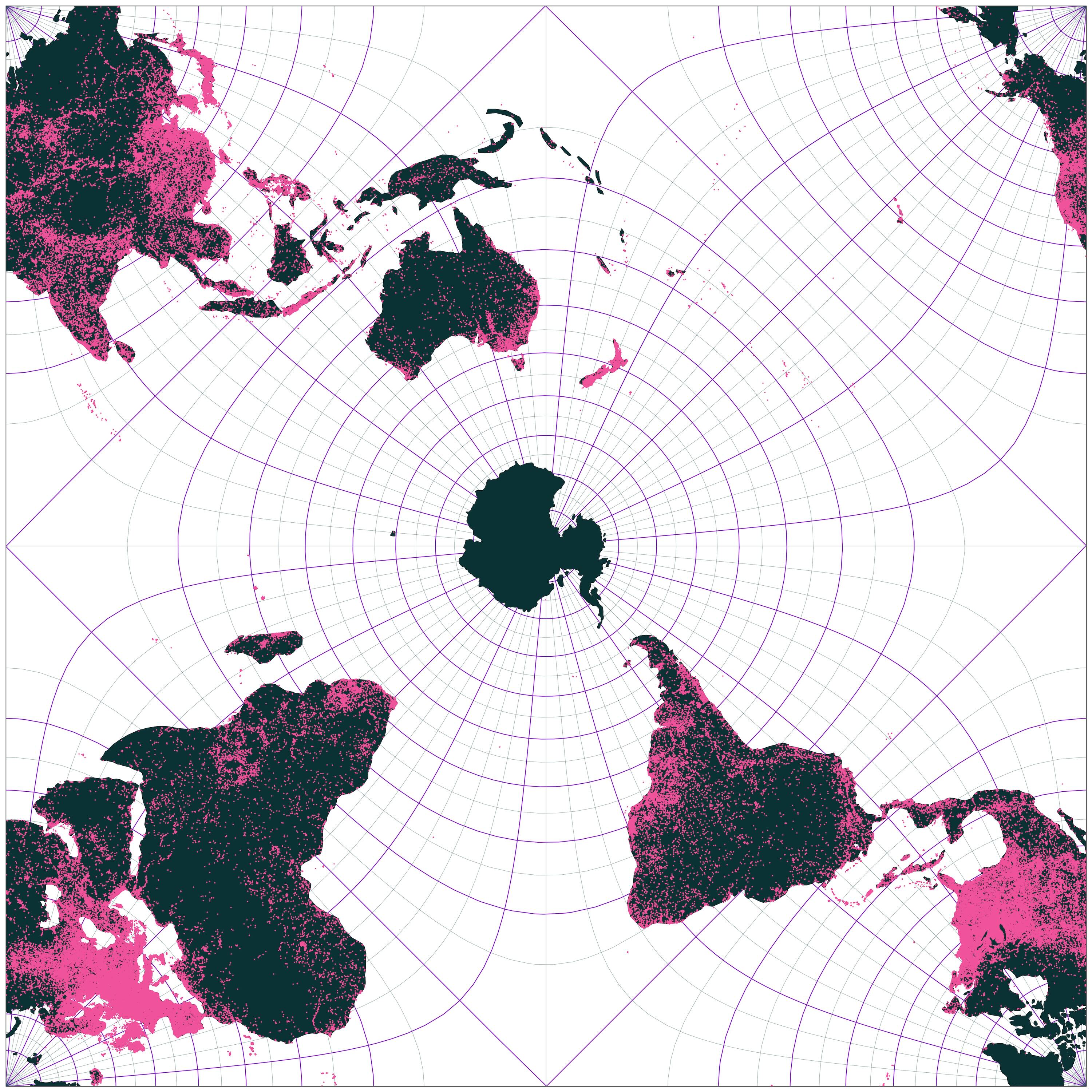}
        };
        \node[inner sep=0, scale=.6, xshift=14.9mm, yshift=3mm] at (current page.center){
        \parbox{50mm}{\captionof{figure}{\raggedright \\
        \hspace{-5mm}They may\\
        \hspace{-6mm}take our \\
        \hspace{-5mm}Introduction, \\
        \hspace{-7.3mm}but they'll never take\\
        our Antarctica!}}
    };
    \end{tikzpicture}
    \end{figure*}
}

\end{document}